\documentclass[lettersize,journal]{IEEEtran}
\IEEEoverridecommandlockouts 

\usepackage{verbatim}
\usepackage{hyperref} 
\usepackage{graphicx, caption}
\usepackage{subcaption}
\usepackage{amssymb}
\usepackage{amsmath}
\usepackage{amsthm}
\usepackage{mathtools}
\usepackage{cite}
\usepackage{stfloats}
\usepackage{epstopdf}
\usepackage{psfrag}
\usepackage[mathscr]{euscript}
\usepackage{acronym}  
\usepackage{booktabs} 
\usepackage[linesnumbered,ruled,vlined]{algorithm2e}
\usepackage[table]{xcolor}

\SetKwInput{KwInput}{Input}
\SetKwInput{KwOutput}{Output}

\usepackage{color}
\usepackage{dsfont}
\usepackage{bbm}

\acrodef{IoT}{Internet of Things}
\acrodef{BS}{base station}
\acrodef{pdf}{probability density function}
\acrodef{i.i.d.}{independent and identically distributed}
\acrodef{CDF}{cumulative distribution function}
\acrodef{FL}{federated learning}
\acrodef{ML}{machine learning}
\acrodef{SGD}{stochastic gradient descent}
\acrodef{SG}{stochastic gradient}
\acrodef{MAC}{multiply-accumulate}
\acrodef{CNN}{convolutional neural network}
\acrodef{DNN}{deep neural network}
\acrodef{QNN}{quantized neural network}
\acrodef{OFDMA}{orthogonal frequency domain multiple access}
\acrodef{FDMA}{frequency domain multiple access}
\acrodef{NBI}{normal boundary inspection}
\acrodef{CHIM}{convex hull of individual minima}
\acrodef{NBS}{Nash bargaining solution}
\acrodef{SUM}{sum minimizing solution}

\usepackage{color}
\usepackage{dsfont}
\usepackage{bbm}

\newtheorem{theorem}{Theorem}
\newtheorem{lemma}{Lemma}

\newtheorem{assumption}{Assumption}

\newcommand{\limit}[2]{\lim_{#1 \to\infty} #2}
\newcommand{\E}{\mathbb{E}}

\newcommand{\bo}[1]{\boldsymbol{#1}}
\newcommand{\Whole}{N}
\newcommand{\weights}{\boldsymbol{w}}
\newcommand{\qweights}{\boldsymbol{w}^{Q, k}}
\newcommand{\inputvec}{\boldsymbol{x}_{kl}}
\newcommand{\outputvec}{\boldsymbol{y}_{kl}}
\newcommand{\datasize}{D_k}
\newcommand{\loss}{F_k}
\newcommand{\globall}{F}
\newcommand{\globalit}{t}
\newcommand{\scheduleset}{\mathcal{N}_t}
\newcommand{\schedulesize}{K}
\newcommand{\SGDrun}{I}
\newcommand{\precision}{n}
\newcommand{\tprecision}{m}
\newcommand{\globalnum}{T}
\newcommand{\smallest}{\kappa}
\newcommand{\Quantize}{Q}
\newcommand{\ka}{\nonumber \\}
\newcommand{\learningrate}{\eta}
\newcommand{\minibatch}{\xi}
\newcommand{\modelupdate}{\boldsymbol{d}}
\newcommand{\qmodelupdate}{\boldsymbol{d}^{Q,k}}
\newcommand{\smooth}{L}
\newcommand{\strong}{\mu}
\newcommand{\sgdvariance}{\sigma}
\newcommand{\sgdbound}{G}

\newcommand{\paramsize}{d}
\newcommand{\MACunits}{p}
\newcommand{\MACE}{E_\text{MAC}}

\newcommand{\mainb}{E_\text{m}(\precision)}
\newcommand{\MACamplitude}{A}
\newcommand{\premax}{n_{\text{max}}}
\newcommand{\tpremax}{m_{\text{max}}}
\newcommand{\MACex}{\alpha}
\newcommand{\localE}{E^{C, k} (n)}
\newcommand{\inferenceE}{E^k_{\text{inf}}(\precision)} 
\newcommand{\updateweightE}{E_\text{back}}
\newcommand{\DRAM}{E_\text{D}}
\newcommand{\DRAME}{E_\text{DRAM}(n)}
\newcommand{\DRAMCO}{A_d}
\newcommand{\tE}{E^{UL, k} (m)}
\newcommand{\numweights}{d}
\newcommand{\numMAC}{\text{$N_c$}}
\newcommand{\numout}{\text{$O_c$}}
\newcommand{\achievalble}{r}
\newcommand{\accuracy}{\epsilon}
\newcommand{\obj}{g}
\newcommand{\objone}{\obj_1(\SGDrun, \schedulesize, \tprecision, \precision)}
\newcommand{\objtwo}{\obj_2(\SGDrun, \schedulesize, \tprecision, \precision)}
\newcommand{\objvec}{\boldsymbol{\obj}}
\newcommand{\globalset}{\mathcal{I}}
\newcommand{\vweights}{\boldsymbol{v}}
\newcommand{\avgv}{\bar{\vweights}}
\newcommand{\avgweights}{\bar{\weights}}
\newcommand{\avggd}{\bar{\delta}}
\newcommand{\avgsgd}{\delta}
\newcommand{\smart}{\rho}
\newcommand{\Noniid}{{\Gamma}}
\newcommand{\uweights}{\boldsymbol{u}}
\newcommand{\avgu}{\bar{\uweights}}
\newcommand{\longconst}{\psi_1}
\newcommand{\longconsttwo}{\psi_2}
\newcommand{\power}{P^{\text{tx}}}
\newcommand{\prob}{p_k}
\newcommand{\payoff}{\Phi}
\newcommand{\penaltyterm}{\lambda}
\newcommand{\disagreement}{\boldsymbol{D}}
\newcommand{\achievable}{\obj_\text{ach}}

\allowdisplaybreaks 

\begin{document}
	\newcommand{\paperTitle}{Title}
	\vspace{-0.0cm}
	
	
	\title{\vspace{-0.0cm} 
		Green, Quantized Federated Learning over Wireless Networks: An Energy-Efficient Design	
}
	\author{ \vspace{-0.0 cm}	
		
		\IEEEauthorblockN{
			Minsu~Kim, Walid~Saad, \textit{Fellow, IEEE}, Mohammad~Mozaffari, \textit{Member IEEE}, and Merouane~Debbah, \textit{Fellow, IEEE}
		}\\[-0.5em]
		\vspace{-0.0mm}
		\thanks{M.\ Kim and W.\ Saad are with the Wireless@VT Group, Bradley Department of Electrical and Computer Engineering, Virginia Tech, Arlington, VA, USA (email: \{\texttt{msukim, walids}\}@vt.edu.)
			
		M.\ Mozaffari is with Ericsson Research, Santa Clara, CA, USA (email: \texttt{mohammad.mozaffari@ericsson.com}).
		
		M.\ Debbah is with Khalifa University of Science and Technology, Abu Dhabi, UAE (email: \texttt{ merouane.debbah@ku.ac.ae}).
		
		A preliminary version of this work was presented at IEEE ICC 2022 \cite{ICC_version}.
		
		This research was supported by the U.S. National Science Foundation under Grant CNS-2114267.
		
		The implementation code is available on https://github.com/news-vt
				}
	}

	\maketitle 
	\vspace{-0.0cm}

	\acresetall
	\begin{abstract}
	 The practical deployment of federated learning (FL) over wireless networks requires balancing energy efficiency, convergence rate, and a target accuracy due to the limited available resources of devices.
	 Prior art on FL often trains deep neural networks (DNNs) to achieve high accuracy and fast convergence using 32 bits of precision level. However, such scenarios will be impractical for resource-constrained devices since DNNs typically have high computational complexity and memory requirements. Thus, there is a need to reduce the precision level in DNNs to reduce the energy expenditure. 
	 In this paper, a green-quantized FL framework, which represents data with a finite precision level in both local training and uplink transmission, is proposed. Here, the finite precision level is captured through the use of quantized neural networks (QNNs) that quantize weights and activations in fixed-precision format. In the considered FL model, each device trains its QNN and transmits a quantized training result to the base station. Energy models for the local training and the transmission with quantization are rigorously derived. To minimize the energy consumption and the number of communication rounds simultaneously, a multi-objective optimization problem is formulated with respect to the number of local iterations, the number of selected devices, and the precision levels for both local training and transmission while ensuring convergence under a target accuracy constraint. To solve this problem, the convergence rate of the proposed FL system is analytically derived with respect to the system control variables. Then, the Pareto boundary of the problem is characterized to provide efficient solutions using the normal boundary inspection method. Design insights on balancing the tradeoff between the two objectives while achieving a target accuracy are drawn from using the Nash bargaining solution and analyzing the derived convergence rate. Simulation results show that the proposed FL framework can reduce energy consumption until convergence by up to 70\% compared to a baseline FL algorithm that represents data with full precision without damaging the convergence rate.
	\end{abstract}

	\section{Introduction}
	\Ac{FL} is an emerging paradigm that enables distributed learning among wireless devices \cite{CM:21}. In \ac{FL}, a central server (e.g., a \ac{BS}) and multiple mobile devices collaborate to train a shared machine learning model without sharing raw data. Many \ac{FL} works employ \acp{DNN}, whose size constantly grows to match the increasing demand for higher accuracy \cite{Ro:19}. Such \ac{DNN} architectures can have tens of millions of parameters and billions of \ac{MAC} operations. Moreover, to achieve fast convergence, these networks typically represent data in 32 bits of full precision level, which may consume significant energy due to high computational complexity and memory requirements \cite{HB:16}. Additionally, a large \ac{DNN} can induce a significant communication overhead \cite{MB:17}. Under such practical constraints, it may be challenging to deploy \ac{FL} over resource-constrained \ac{IoT} devices due to its large energy cost. To design an energy-efficient, green \ac{FL} scheme, one can reduce the precision level to decrease the energy consumption during the local training and communication phase. However, a low precision level can jeopardize the convergence rate by introducing quantization errors. Therefore, finding the optimal precision level that balances energy efficiency and convergence rate while meeting desired FL accuracy constraints will be a major challenge for the practical deployment of green \ac{FL} over wireless networks. 
	
	Several works have studied the energy efficiency of \ac{FL} from a system-level perspective \cite{Stesa:21, Ngutra:19, Zhaya:21, Khaung:21, BL:21, BA:20}. The work in \cite{Stesa:21} investigated the energy efficiency of \ac{FL} algorithms in terms of the carbon footprint compared to centralized learning. In \cite{Ngutra:19}, the authors formulated a joint minimization problem for energy consumption and training time by optimizing heterogeneous computing and wireless resources. The work in \cite{Zhaya:21} developed an approach to minimize the total energy consumption by controlling a target accuracy during local training based on a derived convergence rate. The authors in \cite{Khaung:21} proposed a sum energy minimization problem by considering joint bandwidth and workload allocation of heterogeneous devices. In \cite{BL:21}, the authors studied a joint optimization problem to minimize the energy and the training time under a target accuracy. The work in \cite{BA:20} developed a resource management scheme by leveraging the information of loss functions of each device to maximize the accuracy under constrained communication and computation resources. However, these works \cite{Stesa:21, Ngutra:19, Zhaya:21, Khaung:21, BL:21, BA:20} did not consider the energy efficiency of their \ac{DNN} structure during training. Since mobile devices have limited computing and memory resources, deploying an energy-efficient \ac{DNN} will be necessary for green \ac{FL}.
	
To further improve \ac{FL} energy efficiency, model compression methods such as quantization were studied in \cite{additional1, addtional2, FE:21, Ch:22}. In \cite{additional1}, the authors developed an over-the-air \ac{FL} system that uses one-bit gradient quantization aggregation scheme. The authors in \cite{addtional2} developed an approach to minimize the training time by optimizing transmission precision level and bandwidth allocation. The work in \cite{FE:21} proposed an approach to minimize the energy consumption and the loss function by optimizing model compression design for uplink transmission and device selection strategy. In \cite{Ch:22}, the authors studied an energy minimization problem by controlling local iterations, bandwidth allocation, and precision level for both local training and transmission under full device participation scheme. However, the works in \cite{additional1, addtional2, FE:21} only considered the communication efficiency while there can be a large energy consumption in local training due to high precision level. Although the work in \cite{Ch:22} considered quantization for both local training and transmission, it used full device participation scheme, which is not practical due to stragglers, and only the energy consumption is minimized. In our previous work \cite{ICC_version}, an energy minimization problem was formulated to investigate the tradeoff between energy, precision, and accuracy. However, the same precision level was used for local training and transmission as done in \cite{Ch:22}. As such, the results of \cite{ICC_version} cannot be directly applied for more general cases such as those with heterogeneous devices and non-i.i.d datasets. Moreover, the number of local iterations and the number of selected devices were not jointly optimized. To the best of our knowledge, there are no current works that jointly consider the tradeoff between energy efficiency, convergence rate, and accuracy while simultaneously controlling local iterations, the number of scheduled devices, and precision levels in local training and transmission for green \ac{FL} over wireless networks.

	The main contribution of this paper is a novel green, energy-efficient quantized \ac{FL} framework that can represent data with a finite precision level in both local training and uplink transmission. Our contributions include:
	\begin{itemize}
	\item We propose an FL framework that takes into account stochastic quantization in both local training and transmission with different precision levels. All devices train their \acp{QNN}, whose weights and activations are quantized with a finite precision level, so as to decrease energy consumption for computation and memory access. In uplink communication, each device performs quantization to its training result to improve the communication efficiency.
	
	\item To quantify the energy consumption, we propose a rigorous energy model for the local training based on the physical structure of a processing chip. We also derive the energy model for the uplink transmission with quantization. Although a low precision level can save the energy consumption per iteration, it decreases the convergence rate because of quantization errors. Thus, there is a need for a new approach to analyze the tradeoff between energy efficiency, convergence rate, and target accuracy by optimizing the precision levels. To this end, we formulate a novel multi-objective optimization problem by controlling the precision levels to jointly minimize the total energy consumption and the number of communication rounds while ensuring convergence with a target accuracy. We also incorporate two additional control variables: the number of local iterations and the number of selected devices at each communication round, which have a significant impact on both the energy consumption and the convergence time.

	\item To solve this problem, we first analytically derive the convergence rate of our \ac{FL} framework with respect to the control variables under non-iid data distribution. We then optimize sampling probabilities for devices based on the derived convergence rate. Subsequently, we use the \ac{NBI} method to obtain the Pareto boundary of our multi-objective optimization problem. To balance the tradeoff between the two objectives, we present and analyze two practical operating points: the \ac{NBS} and the \ac{SUM} points.
	
	\item Based on the aforementioned operating points and the derived convergence rate, we provide design insights into the proposed \ac{FL} framework. For instance, the total energy consumption initially decreases as the precision levels increase, however, after a certain threshold, a higher precision will induce higher energy costs. Meanwhile, the convergence rate will always improve with a higher precision. However, this improvement becomes negligible after a certain level. We also show that we need a higher precision level to achieve higher target accuracy at the expense of more energy and communication rounds. We then provide the impacts of system parameters such as the number of devices and model size on the performance of the proposed \ac{FL}.
	\end{itemize}
	Simulation results show that our \ac{FL} model can reduce the energy consumption around 70\% compared to FedAvg without damaging the convergence rate.
		
	The rest of this paper is organized as follows. Section \ref{sec:system model} presents the system model. In Section \ref{sec:problem formulation}, we describe the studied problem. Section \ref{sec:simlulation results} provides simulation results. Finally, conclusions are drawn in Section \ref{sec:conclusion}.

\begin{table*}[t]
	\caption{List of notations.}
	\vspace{-0.0cm}
	\begin{center}
	\scalebox{0.9}{
	\begin{tabular}{|c|c|c|c|} 
		\hline
		Notation                   & Description                                & Notation       & Description                                   \\ \hline
		$\Whole$                   & Number of devices                          & $\power_k$     & Transmission power                            \\ \hline
		$(\inputvec, \outputvec)$ & Data sample                                & $\bar{h}_k$          & Average channel gain                                  \\ \hline
			$\datasize$                & Dataset size                               & $N_0$          & Power spectral density of noise               \\ \hline
			$\weights^k$               & Model parameters                           & $\tE$          & Energy consumption for uplink transmission    \\ \hline
			$\loss(\cdot)$             & Local loss function                        & $\smooth$      & Smoothness parameter                          \\ \hline
			$\SGDrun$                  & Number of local iterations                 & $\strong$      & Convexity parameter                           \\ \hline
			$\schedulesize$            & Number of sampled devices                  & $\Noniid$      & Degree of non-iidness                         \\ \hline
			$\tprecision$              & Precision level for transmission           & $\sgdbound$    & Bound of the norm of stochastic gradients     \\ \hline
			$\precision$               & Precision level for local training         & $\sgdvariance$ & Bound of the variance of stochastic gradients \\ \hline
			$\accuracy$                & Target accuracy                            & $\paramsize$   & Number of model parameters                \\ \hline
			$\localE$                  & Energy consumption for one local iteration & $N_c$          & Number of MAC operations                  \\ \hline
			$B$                        & Allocated bandwidth                        & $O_s$          & Number of neurons                         \\ \hline
		\end{tabular}}
	\end{center}
	\label{tab:notation} \vspace{-0.0cm}
	\end{table*}

	\begin{figure}
		\centering
		\includegraphics[width=0.95\columnwidth]{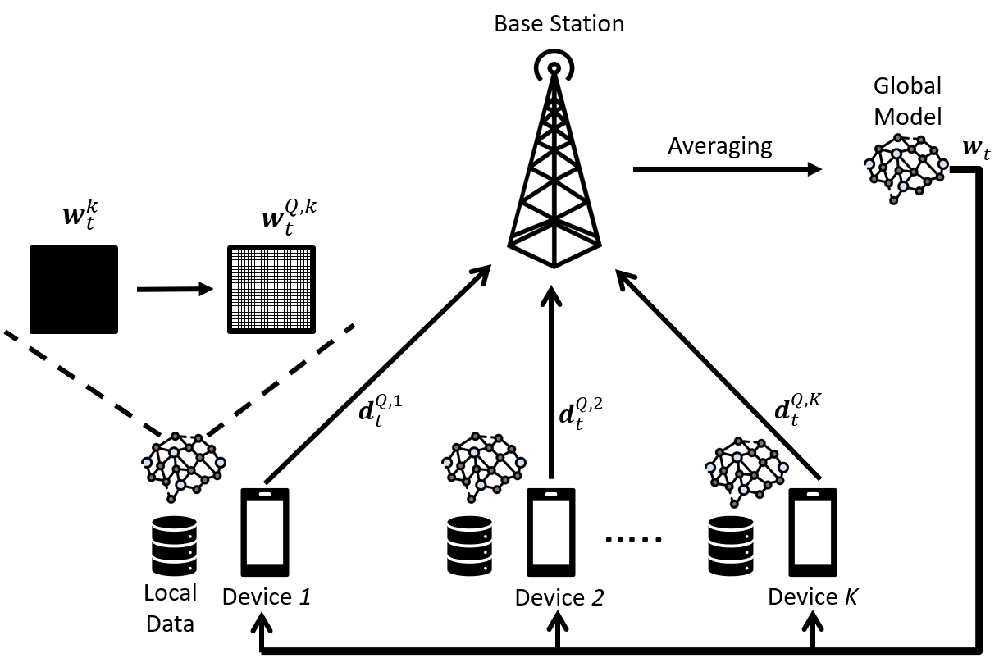}
		\captionsetup {singlelinecheck = false}

		\vspace{-0.0cm}
		\caption{An illustration of the quantized \ac{FL} model over wireless network.}
		\label{fig:system_model}

		\vspace{-0.0 cm}	
	\end{figure}
	
	\section{System Model} \label{sec:system model}
Consider an \ac{FL} system having $\Whole$ devices connected to a \ac{BS} as shown in Fig. \ref{fig:system_model}. Each device $k$ has its own local dataset $\mathcal{D}_k = \{\inputvec, \outputvec\}$, where  $\l = 1, \dots, \datasize$. For example, $\{\inputvec, \outputvec\}$ can be an input-output pair for image classification, where $\boldsymbol{x}_{kl}$ is an input vector and $\outputvec$ is the corresponding output. We define a loss function $f(\weights^k,\inputvec, \outputvec )$ to quantify the performance of a \ac{ML} model with parameters $\weights^k \in \mathbb{R}^d$ over $\{\inputvec, \outputvec\}$, where $d$ is the number of parameters. Since device $k$ has $\datasize$ data samples, its local loss function can be given by $	\loss(\weights^k) = \frac{1}{\datasize} \sum_{l=1}^{\datasize}  f(\weights^k,\inputvec,\outputvec ).$
%
%
%
%
The \ac{FL} process aims to find the global parameters $\weights$ that can solve the following optimization problem:
\begin{align}
	\min_{\weights^1, \dots, \weights^\Whole} & \quad \hspace{-3.0mm} \globall(\weights) \hspace{-0.5mm} =  \hspace{-1.0mm} \sum_{k=1}^{\Whole} \hspace{-0.5mm} \frac{\datasize}{D} \loss (\weights^k) \hspace{-0.5mm} = \hspace{-0.5mm} \frac{1}{D} \hspace{-0.5mm} \sum_{k=1}^{N} \sum_{l=1}^{\datasize} f(\weights^k,\inputvec, \outputvec ) \\
	\textrm{s.t.} \quad & \weights^1 = \weights^2 = \dots = \weights^\Whole =\weights, \label{FL problem}
\end{align}
where $D = \sum_{k=1}^{\Whole} \datasize$ is the total size of the entire dataset $\mathcal{D} = \cup_{k=1}^{\Whole} \mathcal{D}_k$. Without loss of generality, we assume datasets across devices are non-iid.

Solving problem \eqref{FL problem} typically requires an iterative process between the \ac{BS} and devices. However, in practical systems, such as \ac{IoT} systems, these devices are resource-constrained, particularly when it comes to computing and energy. Hence, we propose to manage the precision level of parameters used in our \ac{FL} algorithm to reduce the energy consumption for computation, memory access, and transmission. As such, we adopt a \ac{QNN} architecture whose weights and activations are quantized in fixed-point format rather than conventional 32-bit floating-point format \cite{IH:16}. During the training time, a \ac{QNN} can reduce the energy consumption for \ac{MAC} operation and memory access due to quantized weights and activations.
\subsection{Quantized Neural Networks} \label{QNN}
In our model, each device trains a \ac{QNN} of identical structure using $\precision$ bits for quantization. High precision can be achieved if we increase $\precision$ at the cost of more energy usage. We can represent any given number in a fixed-point format such as $[\Omega. \payoff]$, where $\Omega$ is the integer part and $\payoff$ is the fractional part of the given number \cite{SG:15}. Here, we use one bit to represent the integer part and $(\precision-1)$ bits for the fractional part. Then, the smallest positive number that we can present is $\kappa = 2^{-n+1}$, and the possible range of numbers with $\precision$ bits will be $[-1 ,1-2^{-n+1}]$. Note that a \ac{QNN} restricts the value of weights to [-1, 1]. Otherwise, weights can be very large without meaningful impact on the performance. We consider a stochastic quantization scheme \cite{SG:15} since it generally performs better than deterministic quantization \cite{DOREFA}. Any given number $w \in \weights$ can be stochastically quantized as follows:
\begin{align}
	\Quantize(w)  = 
	\begin{cases}
		\lfloor w \rfloor, & \quad  \text{with probability} \quad \frac{\lfloor w \rfloor +\smallest - w}{\smallest},\\
		\lfloor w \rfloor + \smallest,              & \quad \text{with probability} \quad \frac{w - \lfloor w \rfloor}{\smallest},
	\end{cases}
\end{align}    	
where $\lfloor w \rfloor$ is the largest integer multiple of $\smallest$ less than or equal to $w$. In the following lemma, we analyze the features of the stochastic quantization.
\begin{lemma}
	For the stochastic quantization $\Quantize(\cdot)$, a scalar $w$, and a vector $\weights \in \mathbb{R}^d$, we have \label{Lem1}
	\begin{align}
		&\E[\Quantize(w)] = w, \quad \E[(\Quantize(w) - w)^2] \leq \frac{1}{2^{2\precision}}, \\
		&\E[\Quantize(\weights)] = \weights, \quad \E[||\Quantize(\weights) - \weights||^2] \leq \frac{d}{2^{2\precision}}. \label{quant_variance}
	\end{align}
\end{lemma}
\begin{proof}
	We first derive $\E[\Quantize(w)]$ as
	\begin{align}
		\E[\Quantize(w)] &= \lfloor w \rfloor \frac{\lfloor w \rfloor \hspace{-0.5mm} + \hspace{-0.5mm} \smallest \hspace{-0.5mm} - \hspace{-0.5mm} w}{\smallest} \hspace{-0.5mm} + \hspace{-0.5mm} (\lfloor w \rfloor \hspace{-0.5mm} +  \hspace{-0.5mm} \smallest) \frac{w \hspace{-0.5mm} - \hspace{-0.5mm} \lfloor w \rfloor}{\smallest} \hspace{-0.5mm}  = \hspace{-0.5mm} w . \label{Lem1-1}
	\end{align} 
	Similarly,  $\E[(\Quantize(w)-w)^2]$ can be obtained as
	\begin{align}
		\E \hspace{-0.3mm} [ \hspace{-0.3mm} (\Quantize( \hspace{-0.5mm} w \hspace{-0.5mm}) \hspace{-0.7mm} - \hspace{-0.7mm}w)^2 \hspace{-0.2mm}] & \hspace{-0.5mm}=  \hspace{-0.5mm} ( \hspace{-0.3mm} \lfloor \hspace{-0.5mm} w \hspace{-0.5mm} \rfloor \hspace{-0.7mm} -  \hspace{-0.7mm} w)^2 \frac{\lfloor \hspace{-0.5mm} w \hspace{-0.5mm} \rfloor  \hspace{-0.7mm} +  \hspace{-0.7mm} \smallest \hspace{-0.5mm}  - \hspace{-0.5mm}	 w}{\smallest} \hspace{-0.5mm} + \hspace{-0.5mm} (\lfloor\hspace{-0.5mm} w \hspace{-0.5mm} \rfloor \hspace{-0.5mm} + \hspace{-0.5mm} \smallest \hspace{-0.5mm} - \hspace{-0.5mm}w)^2 \frac{w \hspace{-0.5mm} - \hspace{-0.5mm} \lfloor \hspace{-0.5mm} w \hspace{-0.5mm} \rfloor}{\smallest} \ka
		&= (w - \lfloor \hspace{-0.5mm} w \hspace{-0.5mm} \rfloor) (\lfloor w \rfloor+\smallest -w) \leq \frac{\smallest^2}{4} = \frac{1}{2^{2\precision}},  \label{AMGM}
	\end{align}
	where \eqref{AMGM} follows from the arithmetic mean and geometric mean inequality. Since expectation is a linear operator, we have $\E[\Quantize(\weights)] = \weights$ from \eqref{Lem1-1}. From the definition of the square norm, $\E[||\Quantize(\weights) - \weights||^2]$ can obtained as
	\vspace{-0mm}
	\begin{align}
		\E[||\Quantize(\weights) - \weights||^2] = \sum_{j=1}^{d} \E[(\Quantize(w_j) - w_j)^2] \leq \frac{d}{2^{2\precision}}.
	\end{align}
	\vspace{-0.0mm}
\end{proof}
From Lemma \ref{Lem1}, we can see that our quantization scheme is unbiased as its expectation is zero. However, the quantization error can still increase for a large model. 

For device $k$, we denote the quantized weights of layer $l$ as $\qweights_{(l)} = Q(\weights^k_{(l)})$, where $\weights^k_{(l)}$ is the parameters of layer $l$. Then, the output of layer $l$ will be: 
%
%
$o_{(l)} = g_{(l)} (\qweights_{(l)}, o_{(l-1)}),$ where $o_{(l-1)}$ is the output from the previous layer $l-1$, and $g(\cdot)$ is the operation of layer $l$ on the input, including the linear sum of $\qweights_{(l)}$ and  $o_{(l-1)}$, batch normalization, and activation. 
Note that our activation includes the stochastic quantization after a normal activation function such as ReLU. Then, the output of layer $l$, i.e., $o_{(l)}$, is fed into the next layer as an input. For training, we use the \ac{SGD} algorithm as follows
\begin{align}
	\weights^k_{\tau+1} \leftarrow \weights^k_\tau - \learningrate \nabla \loss(\qweights_\tau, \minibatch^k_\tau), \label{SGD}
\end{align}
where $\tau = 1 \dots \SGDrun$ is training iteration, $\learningrate$ is the learning rate, and $\minibatch$ is a sample from $\datasize$ for the current update. The update of weights is done in full precision so that \ac{SG} noise can be averaged out properly \cite{IH:16}. Then, we restrict the values of $\weights^k_{\tau+1}$ to $[-1, 1]$ as
%
%
%
%
$\weights^k_{\tau+1} \leftarrow \text{clip}(\weights^k_{\tau+1}, -1, 1)$
where $\text{clip}(\cdot, -1, 1)$ projects an input to 1 if it is larger than 1, and projects an input to -1 if it is smaller than -1. Otherwise, it returns the same value as the input. Otherwise, $\weights^k_{\tau+1}$ can become significantly large without a meaningful impact on quantization \cite{IH:16}. After each training, $\weights^k_{\tau+1}$ will be quantized as $\qweights_{\tau+1}$ for the forward propagation.  

\subsection{\ac{FL} model}

For learning, without loss of generality, we adopt FedAvg \cite{HB:16} to solve problem \eqref{FL problem}. At each communication round $\globalit$, the \ac{BS} selects $\schedulesize$ devices according to probability $\prob$ for device $k$ such that $\sum_{k=1}^{\Whole} \prob = 1$, and we denote the sampled set as $\scheduleset$. The \ac{BS} transmits the current global model $\weights_t$ to the scheduled devices. Each device in $\scheduleset$ trains its local model based on the received global model by running $\SGDrun$ steps of \ac{SGD} as below
\begin{align}
	\weights_{t, \tau}^k \hspace{-0.5mm}= \hspace{-0.5mm} \weights_{t, \tau-1}^k  \hspace{-0.5mm}- \hspace{-0.5mm} \learningrate_t \nabla \loss(\qweights_{t, \tau-1}, \minibatch^k_\tau), \forall \tau \hspace{-0.5mm} = \hspace{-0.5mm} 1, \dots , \SGDrun , \label{SGD_update}
\end{align}	
where $\learningrate_t$ is the learning rate at communication round $t$.
Note that unscheduled devices do not perform local training. Then, each device in $\scheduleset$ calculates the model update $\modelupdate^k_{t+1} = \weights^k_{t+1} - \weights^k_t$, where $\weights^k_{t+1} = \weights^k_{t, \SGDrun-1}$ and $\weights^k_t = \weights^k_{t, 0}$ \cite{SZ:21}. Typically, $\modelupdate^k_{t+1}$ has millions of elements for \ac{DNN}. It is not practical to send $\modelupdate^k_{t+1}$ with full precision for energy-constrained devices. Hence, we apply the same quantization scheme used in \acp{QNN} to $\modelupdate^k_{t+1}$ by denoting its quantized equivalent as $\qmodelupdate_{t+1}$ with precision level $\tprecision$. Thus, each device in $\scheduleset$ clips its model update $\modelupdate^k_{t+1}$ using $\text{clip}(\cdot)$ to match the quantization range and transmits its quantized version to the \ac{BS}. The received model updates are averaged by the \ac{BS}, and the next global model $\weights_{t+1}$ will be generated as below
\begin{align}
	\weights_{t+1} = \weights_t + \frac{1}{K}\sum_{k \in \mathcal{N}_{t+1} } \qmodelupdate_{t+1}.
\end{align}
The \ac{FL} system repeats this process until the global loss function converges to a target accuracy constraint $\accuracy$.  We summarize this algorithm in Algorithm 1. Next, we propose the energy model for the computation and the transmission of our \ac{FL} system.

	

\begin{algorithm}[t!]
\caption{Quantized \ac{FL} Algorithm} 
\KwInput{$\schedulesize$, $\SGDrun$, initial model $\weights_0$, $t=0$, target accuracy $\accuracy$}
\Repeat{target accuracy $\accuracy$ is satisfied}{
The \ac{BS} randomly selects a subset of devices $\scheduleset$ and transmits $\weights_t$ to the selected devices; \\
Each device $k\in \scheduleset$ trains its \ac{QNN}  by running $\SGDrun$ steps of \ac{SGD} as \eqref{SGD}; \\
Each device $k\in \scheduleset$ transmits $\qmodelupdate_{t+1}$ to the \ac{BS}; \\
The \ac{BS} generates a new global model 
$\weights_{t+1} = \weights_t + \frac{1}{K}\sum_{k \in \scheduleset} \qmodelupdate_{t+1}$; \\
$t \leftarrow t+1$;

} 
\end{algorithm} 

\subsection{Computing and Transmission model} \label{sec:computing_model}
\subsubsection{Computing model} 
\begin{figure}
	\begin{center}
	\includegraphics[width=1.0\columnwidth]{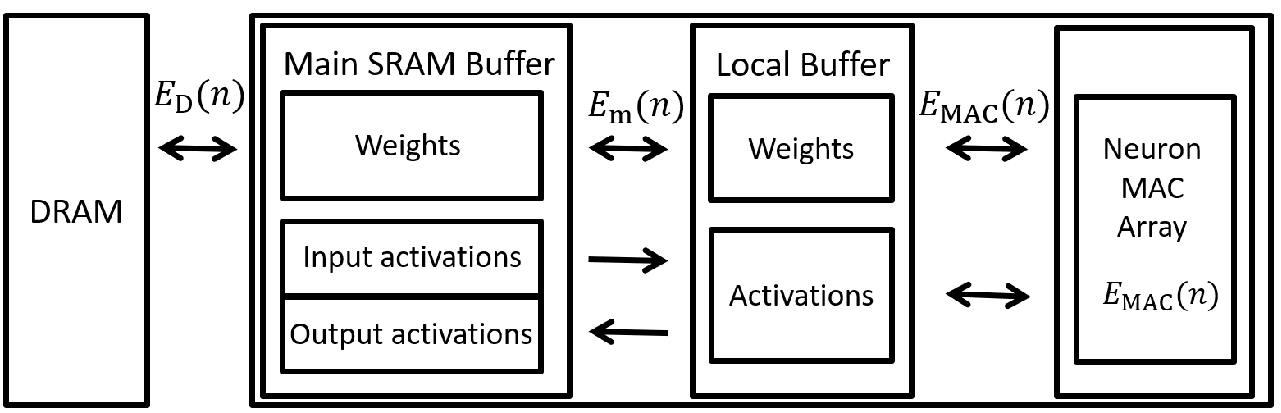}
	\captionsetup {singlelinecheck = false}
	
	\vspace{-0.0cm}
	
	\caption{An illustration of the two-dimensional processing chip.}
	\label{fig:comptuing_model}
	
	\vspace{-0.3cm}
	\end{center}
\end{figure}
We consider a typical two-dimensional processing chip for \acp{CNN} as shown in Fig. 2 \cite{MB:17}. This chip has a DRAM, a parallel neuron array with $\MACunits$ \ac{MAC} units, and two memory levels: a main SRAM buffer that stores the weights and activations and a local buffer that caches currently used weights and activations. Since the main SRAM buffer has a limited size, the input dataset is stored in the DRAM. Some weights can also be stored in the DRAM, if the whole weights cannot be fit in the main SRAM buffer. We use the \ac{MAC} operation energy model of \cite{MB:18} whereby $\MACE (\precision) = \MACamplitude \left( \precision /\premax \right) ^{\MACex}$ for precision level $\precision$, where $\MACamplitude > 0$, $1 < \MACex < 2$, and $\premax$ is the maximum precision level. Here, a \ac{MAC} operation includes neuron output calculation, batch normalization, activation, and back-propagation. From \cite{MB:18}, the energy consumption for accessing a local buffer can be modeled as $\MACE (\precision)$, and the energy for accessing a main buffer can be given by $\mainb = 2 \MACE (\precision)$. The energy consumption to access a DRAM can be modeled as $\DRAM(\precision) = \DRAMCO \MACE (\precision)$, where $\DRAMCO >> 1$. \cite{MB:17}.

The energy consumption of device $k$ for doing inference (i.e., forward propagation) is $\inferenceE$ when $\precision$ bits are used for the quantization. Then, $\inferenceE$ is the sum of the computing energy $E_\text{C}(\precision)$, the access energy for fetching weights from the buffers  $E_\text{W}(\precision)$, the access energy for fetching activations from the buffers $E_\text{A}(\precision)$ and the access energy for fetching input features and weights from the DRAM $\DRAME$, as follows \cite{MB:18}:
\begin{align}
	&\inferenceE = E_\text{C}(\precision) + E_\text{W}(\precision) + E_\text{A}(\precision) + \DRAME, \ka
	&E_{\text{C}}(\precision) = \MACE (\precision) \numMAC + 2\numout \ \MACE(\premax), \ka &E_\text{W}(\precision) = \mainb \numweights + \MACE (\precision) \numMAC \sqrt{{\precision}/{\MACunits \premax}}, \ka
	&E_\text{A}(\precision) = 2 \mainb \numout +  \MACE (\precision) \numMAC \sqrt{{\precision}/{\MACunits \premax}}, \ka
	&\DRAME = \DRAM(\premax) x_\text{in} + 2\DRAM(\precision) \max(\paramsize\precision + \numout \precision - S, 0) ,\label{computation_energy}
\end{align} 
where $\numMAC$ is the number of \ac{MAC} operations, $\numweights$ is the number of weights, $\numout$ is the number of intermediate outputs in the network, $x_\text{in}$ is the input dimension, and $S$ is the size of the main SRAM buffer. For $E_{\text{C}}(\precision)$, in a \ac{QNN}, batch normalization and activation are done in full-precision $\premax$ to each output \cite{IH:16}. We store quantized weights and activations in the SRAM main buffer. Once we fetch weights from a main to a local buffer, they can be reused in the local buffer afterward as shown in $E_\text{W}(\precision)$. In Fig. \ref{fig:comptuing_model}, a \ac{MAC} unit fetches weights from a local buffer to do computation. Since we are using a two-dimensional \ac{MAC} array of $\MACunits$ \ac{MAC} units, they can share fetched weights with the same row and column, which has $\sqrt{\MACunits}$ \ac{MAC} units respectively. In addition, a \ac{MAC} unit can fetch more weights due to the $\precision$ bits quantization compared with when weights are represented in $\premax$ bits. Thus, we can reduce the energy consumption to access a local buffer by the amount of $\sqrt{{\precision}/{\MACunits \premax}}$. A similar process applies to $E_\text{A}(\precision)$ since activations are fetched from the main buffer and should be saved back to it for the calculation in the next layer. For $\DRAME$, input features are processed in full-precision, and weights that cannot be stored in the SRAM will be fetched and stored to the DRAM.

As introduced in Section \ref{QNN}, we calculate gradients in full-precision to average out the noise from \ac{SGD}. In back-propagation, each layer calculates the gradients of its weights and the gradients of the activations of the previous layer. Hence, we can approximate the number of MAC operations as $2\numMAC$ as done in \cite{Ev:20}. Then, the energy consumption for back-propagation is
\begin{align}
	\updateweightE &= 2\numMAC \MACE(\premax) + 2E_\text{m}(\premax)\numout + E_\text{m}(\premax)\paramsize \ka
	&\quad + 2\MACE(\premax) \numMAC \sqrt{\frac{1}{p}} \ka
	& \quad +2  \DRAM(\precision_\text{max}) \max(d\precision_\text{max}
	+ O_c\precision_\text{max} -s_m , 0).
\end{align}
Since back-propagation is done in full-precision, weights  must first be fetched from the DRAM. Then, we fetch weights from the main buffer to the local buffer. The neuron \ac{MAC} array proceeds with the calculation by fetching the cached weights and activations from the local buffer. Therefore, the energy consumption for one iteration of device $k$ is given by
\begin{align}
	\localE = \inferenceE + \updateweightE, \ k \in \{1, \dots, \Whole\}.
\end{align}
\subsubsection{Transmission Model}
We use the \ac{OFDMA} to transmit model updates to the \ac{BS}. Each device occupies one resource block. The achievable rate of device $k$ will be:
\begin{align}
	\achievalble_k = B \log_2\left( 1 +\frac{\power_k \bar{h}_k}{N_0 B}  \right),
\end{align} 
where $B$ is the allocated bandwidth, $\bar{h}_k$ is the average channel gain between device $k$ and the \ac{BS} during training\footnote{For future work, our approach can be extended to the case with instantaneous time-varying channels by considering a stochastic optimization formulation.}\footnote{An important subject of future work here can be the integration of more advanced MIMO-based communication channels.}, $\power_k$ is the transmit power of device $k$, and $N_0$ is the power spectral density of white noise. After local training, device $k$ normalizes the model update as $\modelupdate^k_t / ||\modelupdate^k_t||$ to match the predetermined quantization range $[-1, 1]$. Then, it transmits $\qmodelupdate_t$ to the \ac{BS} at given communication round $t$. 
The transmission time $T_k$ for uploading $\qmodelupdate_t$ is given by
\begin{align}
	\displaystyle T_k (\tprecision) = \frac{\paramsize \tprecision}{\achievalble_k}.
\end{align}
Then, the energy consumption for the uplink transmission is given by
\begin{align}
	\tE = T_k(\tprecision) \times \power_k = \frac{\power_k \paramsize \tprecision} {B \log_2\left( 1 +\frac{ \power_k \bar{h}_k}{N_0 B}  \right)} . \label{transmission_energy}
\end{align}

	
	\section{Time and Energy efficient Federated \ac{QNN}} \label{sec:problem formulation}
Given our model, we now formulate a multi-objective optimization problem to minimize the energy consumption and the number of communication rounds while ensuring convergence under a target accuracy. We show that a tradeoff exists between the energy consumption and the number of communication rounds as a function of $\SGDrun$, $\schedulesize$, $\tprecision$, and $\precision$. For instance, we can reduce the amount of energy spent per iteration by using low precision and sampling a small number of devices. However, this slows the convergence rate because of quantization errors. Meanwhile, the system can allocate more bits and sample more devices to converge faster, i.e, to reduce the number of communication rounds, at the expense of spending more energy. However, this improvement becomes negligible after a certain threshold as shown later in the convergence analysis and simulations (see Theorem \ref{Thm1} and Section \ref{sec:simlulation results}. Hence, finding the optimal solutions is important to balance this tradeoff and to achieve the target accuracy.

We aim to minimize both the expected total energy consumption and the number of communication rounds\footnote{Minimizing the total training time by considering the impact of quantization on the computation time can be an important subject of future research.}  until convergence  under a target accuracy $\accuracy$ as follows:
\begin{subequations}
	\begin{align}
		\min_{\SGDrun, \schedulesize, \tprecision, \precision} & \quad \left[\E \left[ \sum_{t=1}^{\globalnum} \sum_{k\in\scheduleset} \tE + \SGDrun \localE \right], \globalnum  \right] \label{obj} \\
		\textrm{s.t.} \quad & \quad \SGDrun \in [\SGDrun_{\text{min}}, \dots, \SGDrun_{\text{max}}],	\schedulesize \in [\schedulesize_{\text{min}}, \dots, \Whole] \label{I and K constraints} \\ 
		& \quad \tprecision \in [1, \dots, \tpremax],		
		\precision \in [1, \dots, \premax] \label{m and n constraints}\\
		& \quad \E[\globall(\weights_T)] - \globall(\weights^*) \leq \accuracy \label{accuracy_constraint},  
	\end{align}
\end{subequations}
where $\SGDrun$ is the number of local iterations, $\SGDrun_{\text{min}}$ and $\SGDrun_{\text{max}}$ denote the minimum and maximum of $\SGDrun$, respectively, $\E[\globall(\weights_T)]$ is the expectation of global loss function after $T$ communication rounds, $\globall(\weights^*)$ is the minimum value of $\globall$, and $\accuracy$ is the target accuracy. The possible values of $\SGDrun$ and $\schedulesize$ are given by \eqref{I and K constraints}. Constraint \eqref{m and n constraints} represents the maximum precision levels in the transmission and the computation, respectively. Constraints \eqref{accuracy_constraint} captures the required number of communication rounds to achieve $\accuracy$. 

This problem is challenging since the analytical expression of (18d) with respect to the control variables is unknown. Hence, it is not trivial to derive the exact number of $T$ to satisfy (18d). Quantization errors from local training and transmission will slow the convergence rate, thereby making achieving the target accuracy challenging. The convergence is also not always guaranteed under non-iid data distribution. Lastly, a global optimal solution, which minimizes each objective function simultaneously, is generally infeasible for a multi-objective optimization problem \cite{BJ:14}. Therefore, a closed-form solution may not exist due to the tradeoff between two objectives. 

To solve this problem, we first obtain the analytical relationship between \eqref{accuracy_constraint} and $\SGDrun, \schedulesize, \tprecision$, and $\precision$ to derive $T$ with respect to $\accuracy$. As done in \cite{Xi:20, SZ:21, BL:21}, we make the following assumptions on the loss function as follows
\begin{assumption}{The loss function has the following properties} 
	\begin{itemize}
		\item $\loss(\weights)$ is $\smooth$-smooth: $\forall$ $\boldsymbol{v}$ and $\weights$ $\globall_k(\boldsymbol{v}) \leq \globall_k(\weights) + (\boldsymbol{v} - \weights)^T \nabla \globall_k(\weights) + \frac{\smooth}{2} || \boldsymbol{v} - \weights||^2$
		\item $\loss(\weights)$ is $\strong$-strongly convex: $\forall$ $\boldsymbol{v}$ and $\weights$ $\globall_k(\boldsymbol{v}) \geq \globall_k(\weights) + (\boldsymbol{v} - \weights)^T \nabla \globall_k(\weights) + \frac{\strong}{2} || \boldsymbol{v} - \weights||^2$
		\item The variance of \ac{SG} is bounded: $\E[|| \nabla \loss(\weights^k_t, \minibatch^k_t) - \nabla \loss(\weights^k_t)||^2 ] \leq \sgdvariance_k^2, \ \forall k = 1, \dots , N.$
		\item The squared norm of \ac{SG} is bounded: $\E[||\nabla \loss(\weights^k_t, \minibatch^k_t)||^2] \leq \sgdbound^2, \ \forall k= 1, \dots , N$.
	\end{itemize} \label{assumption 1}
\end{assumption}
These assumptions hold for some practical loss functions. Such examples include logistic regression, $l_2$ norm regularized linear regression, and softmax classifier \cite{ZY:13}.
Since we use the quantization in both local training and transmission, the quantization error negatively affects the accuracy and the convergence of our \ac{FL} system. 
We next leverage the results of Lemma \ref{Lem1} so as to derive $T$ with respect to $\accuracy$ in the following theorem. 
\begin{theorem}
	For learning rate $\learningrate_t = \min(\frac{\beta}{t+\gamma}, \frac{1}{\smart}), \beta > \frac{1}{\strong}, \smart \gg 1, \gamma \geq 0$ and, the degree of non-iid $\Noniid = \sum_{k=1}^{\Whole} p_k(\globall_k(\weights^*) - \globall_k^*))$,  we have 
	\begin{align}
		\E[\globall(\weights_T) - \globall(\weights^*)] \leq \frac{\smooth\beta}{2(\beta\strong - 1)}
		\left[
		\frac{\beta \longconsttwo}{T\SGDrun+ \gamma} + \longconst
		\right]
		, \label{T_accuracy}
	\end{align}
	where $\longconst$ and $\longconsttwo$ are 
	\begin{align}
	&\longconst = \frac{\paramsize(\smart - \strong)}{2^{2\precision}}, \ka
	&\longconsttwo = \sum_{k=1}^{\Whole} \prob^2 \sgdvariance_k^2 + 4(\SGDrun-1)^2\sgdbound^2 + \frac{4\paramsize\SGDrun\sgdbound^2}{\schedulesize 2^{2\tprecision}} + \frac{4\SGDrun^2 \sgdbound^2}{\schedulesize} + 4 \smooth \Noniid.
	\end{align} \label{Thm1}
\end{theorem}
\begin{proof}
	See Appendix \ref{proof_Thm1}.
\end{proof}
We can see that $\longconst$ is unavoidable because of the quantization in local training. We also observe that high precision levels for $\precision$ and $\tprecision$ can improve the convergence rate. In particular, we can decrease the quantization error related terms in $\longconst$ and $\longconsttwo$ by increasing $\precision$ and $\tprecision$. However, this improvement becomes negligible after a certain level since those terms decrease exponentially with respect to precision levels.  For $\Noniid$, it quantifies the difference between the loss function at the global optimum $\globall_k(\weights^*)$ and the one at the local optimum $\globall_k^*$. Hence, we can see that the degree of non-iid $\Noniid$ degrades the convergence rate. If we set $\precision = \premax$ and $\tprecision = \tpremax$, we can approximately recover the result of \cite{Xi:20} since the quantization error decays exponentially with respect to $\precision$ and $\tprecision$. The convergence rate also increases with $\schedulesize$. However, all these improvements come at the cost of consuming more energy. We can also see that \eqref{T_accuracy} has the sampling probabilities related term $\sum_{k=1}^{\Whole} \prob^2 \sgdvariance_k^2$ in its numerator. Therefore, we can further improve the convergence rate by optimizing $\prob$ as follows
\begin{align}
	&\min_{p_1, ..., p_\Whole} \quad \sum_{k=1}^{\Whole} \prob^2 \sgdvariance_k^2, \quad \text{s.t.} \quad \sum_{k=1}^{\Whole} \prob = 1, \prob \geq 0, \forall k.
\end{align}
Since the above problem is convex, we can use KKT condition to solve the problem. Then, the optimal sampling probabilities can be given by $\prob = \frac{1/\sgdvariance_k^2}{\sum_{k=1}^{\Whole} 1/\sgdvariance_k^2}.$

From Theorem \ref{Thm1}, we can bound \eqref{T_accuracy} using $\accuracy$ in \eqref{accuracy_constraint} as follows
\begin{align}
	\E[\globall(\weights_\globalnum) - \globall(\weights^*)] \leq \frac{\smooth\beta}{2(\beta\strong - 1)}
	\left[
	\frac{\beta \longconsttwo}{T\SGDrun+ \gamma} + \longconst
	\right] \leq \accuracy. \label{accuracy_inequality}
\end{align}
Since $\longconst$ term is not decreasing with $T$, there exists the minimum value of precision level $\precision_\text{min}$ to achieve $\accuracy$ as follows 
\begin{align}
	\precision_\text{min} = \bigg\lceil \frac{1}{2} \log_2
	\left(
	\smooth\beta \frac{\paramsize(\smart - \strong)}{2\accuracy}
	\right) \bigg\rceil,
\end{align}
where $\lceil \cdot \rceil$ is the smallest integer larger than or equal to the input. To guarantee the convergence, we change the constraint of $\precision$ in \eqref{m and n constraints} as $\precision \in [\precision_\text{min}, \dots, \precision_\text{max}]$. Now, we express each objective function as function of the control variables using Theorem \ref{Thm1}. For notational simplicity, we use $\objone$ for the expected total energy consumption and $\objtwo$ for the number of communication rounds $\globalnum$. Since each device $k$ is selected with probability $\prob$, $\forall k$, we can derive the expectation of the energy consumption in \eqref{obj} as follows
\begin{align}
\objone \hspace{-0.5mm} &=  \hspace{-0.5mm} \E \hspace{-0.5mm}
\left[ \hspace{-0.5mm} \sum_{t=1}^{\globalnum} \sum_{k\in\scheduleset} \tE\hspace{-0.5mm} +\hspace{-0.5mm} \SGDrun \localE \hspace{-0.5mm}
\right] \hspace{-0.5mm} \ka
& = \hspace{-0.5mm} \schedulesize \globalnum \hspace{-0.5mm} \sum_{k=1}^{N} \hspace{-0.5mm} \prob \hspace{-0.5mm} 
\left\{
\tE\hspace{-0.5mm} +\hspace{-0.5mm} \SGDrun \localE
\right\} \label{Energy_formula}. 
\end{align}
Next, we derive $\objtwo$ in a closed-form to fully express the objective functions and to remove the accuracy constraint \eqref{accuracy_constraint}. For any feasible solution that satisfies \eqref{accuracy_constraint} with equality, we can always choose $\globalnum_0 > \globalnum$ such that $\globalnum_0$ still satisfies \eqref{accuracy_constraint}. Since such $\globalnum_0$ will increase the value of the objectives, the accuracy constraint \eqref{accuracy_constraint} should be satisfied with equality \cite{BL:21}. Hence, we take equality in \eqref{accuracy_inequality} to obtain: 
\begin{align}
\objtwo \hspace{-0.7mm}
&= \frac{\beta^2 \longconsttwo} {\SGDrun(\beta\strong-1) (\frac{2\accuracy}{\smooth} - \frac{\beta \longconst}{\beta\strong-1} ) } - \frac{\gamma}{\SGDrun}. \label{Time_formula}
\end{align}
Then, we can change the original problem as below
\begin{subequations}
	\begin{align}
		\min_{\SGDrun, \schedulesize, \tprecision, \precision}  \left[ \objone, \objtwo  \right]   \label{obj_approximation}  \quad	\textrm{s.t.} \quad \eqref{I and K constraints}, \eqref{m and n constraints}.
	\end{align}
\end{subequations}

Since we have two conflicting objective functions, it is infeasible to find a global optimal solution to minimize each objective function simultaneously. Although introducing a weighted sum of the objective functions might provide a unique solution, its optimality is not always guaranteed. We also need to solve the problem again if those weights change. Hence, we instead consider the set of \emph{Pareto optimal points} to obtain an efficient collection of solutions to minimize each objective function and capture the tradeoff. It is known that the set of all Pareto optimal points forms a Pareto boundary in two-dimensional space. Therefore, we use the so-called normal boundary inspection (NBI) method since it provides evenly distributed Pareto optimal points \cite{NBI}.  

We first introduce some terminologies to facilitate the analysis. For a multi-objective function $\objvec(\boldsymbol{x}) = [\obj_1(\boldsymbol{x}), \obj_2(\boldsymbol{x}), \dots \obj_M(\boldsymbol{x})]^T$ and a feasible set $\mathcal{C}$, we define $\boldsymbol{x}_i^*$ as a global solution to minimize $\obj_i(\boldsymbol{x})$, $i = 1 \dots M$, over $\boldsymbol{x} \in \mathcal{C}$. Let $\objvec_i^*=\objvec(\boldsymbol{x}_i^*)$ for $i = 1 \dots M$, and we define the utopia point $\objvec^*$, which is composed of individual global minima $\objvec_i^*$. We define the $M\times M$ matrix $\payoff$, whose $i$th column is $\objvec_i^* - \objvec^*.$ The set of the convex combinations of $\objvec_i^* - \objvec^*$ such that $\{\payoff \boldsymbol{\zeta} \ | \ \zeta_i \geq 0 \ \text{and} \ \sum_{i=1}^{M} \zeta_i = 1 \}$ is defined as \ac{CHIM} \cite{NBI}. For simplicity, we now use $\mathcal{C}$ to represent all feasible constraint sets \eqref{I and K constraints} - \eqref{m and n constraints}. We also define $\boldsymbol{x}_i^*$ as $(\SGDrun, \schedulesize, \tprecision, \precision)$ such that $\obj_i(\SGDrun, \schedulesize, \tprecision, \precision)$ can be minimized over $\mathcal{C}$ for $i = 1$ and $2$.

The basic premise of \ac{NBI} is that any intersection points between the boundary of  $\{\objvec (\SGDrun, \schedulesize, \tprecision, \precision)$ $|$ $(\SGDrun, \schedulesize, \tprecision, \precision) \in \mathcal{C} \}$ and a vector pointing toward the utopia point emanating from the \ac{CHIM} are Pareto optimal. We can imagine that the set of Pareto optimal points will form a curve connecting $\objvec(\boldsymbol{x}_1^*) =[\obj_1(\boldsymbol{x}_1^*), \obj_2(\boldsymbol{x}_1^*)] $ and $\objvec(\boldsymbol{x}_2^*) = [\obj_1(\boldsymbol{x}_2^*), \obj_2(\boldsymbol{x}_2^*)]$. Hence, we first need to obtain $\boldsymbol{x}_1^*$ and $\boldsymbol{x}_2^*$. In the next two subsections, we will minimize $\objone $ and $\objtwo$ separately. 

\subsection{Minimizing $\objone$} \label{sec:minimizing g_1}

Since $\boldsymbol{x}_1^*$ is a global solution to minimize $\objone$, we can find it solving:
\begin{subequations}
	\begin{align}
		\min_{\SGDrun, \schedulesize, \tprecision, \precision} \quad \objone
		  \label{obj1}, \quad
		\textrm{s.t.} \quad (\SGDrun, \schedulesize, \tprecision, \precision) \in \mathcal{C}.
	\end{align}
\end{subequations}
This problem is non-convex because the control variables are an integer and the constraints are not a convex set. For tractability, we relax the control variables as continuous variables. The relaxed variables will be rounded back to integers for feasibility. From \eqref{Energy_formula} and \eqref{Time_formula}, we can see that $\objone$ is a linear function with respect to $\schedulesize$. Therefore, $\schedulesize_{\text{min}}$ always minimizes $\objone$. Moreover, the relaxed problem is convex with respect to $\SGDrun$ since $ \frac{\partial^2 \objone}{\partial \SGDrun^2} > 0$. Hence, we can obtain the optimal $\SGDrun$ to minimize $\objone$ from the first derivative test as
\begin{align}
	\frac{\partial \objone}{\partial \SGDrun} &= H_1 \SGDrun^3 + H_2 \SGDrun^2 + H_3 = 0,
\end{align} 
where
\begin{align}
H_1 &= (8\schedulesize\sgdbound^2 + 8\sgdbound^2) \sum_{k=1}^{\Whole} \prob \localE, \\
H_2 & \hspace{-0.5mm} = (4\schedulesize\sgdbound^2 + 4\sgdbound^2)\sum_{k=1}^{\Whole} \prob \tE  \ka 
&\quad + (-8\schedulesize\sgdbound^2 + \frac{4\paramsize\sgdbound^2}{2^{2\tprecision}}) \sum_{k=1}^{\Whole} \prob \localE, \\
H_3 &= -\schedulesize
\left(
\sum_{k=1}^{\Whole} \prob^2 \sgdvariance_k^2 + 4\sgdbound^2 + 4\smooth \Noniid \right. \ka
&\quad \left. - \gamma (\beta\strong -1)
\left(
\frac{2\accuracy}{\smooth} -\frac{\beta \longconst}{\beta\strong - 1}
\right)
\frac{1}{\beta^2}
\right)
\sum_{k=1}^{\Whole} \prob \tE
\end{align}
Here, $H_1$ and $H_3$ express the cost of local training and the cost of transmission, respectively, while $H_2$ depends on both of them. We next present a closed-form solution of the above equation from Cardano's formula \cite{Cardano}.
\begin{lemma} \label{lem:E_I}
	For given $\tprecision$ and $\precision$, the optimal $\SGDrun'$ to minimize $\objone$ is given by
	\begin{align}
	\SGDrun' & \hspace{-0.5mm} = \hspace{-0.5mm} \sqrt[3]{-\frac{H_2^3}{27H_1^3} \hspace{-0.5mm} - \hspace{-0.5mm} \frac{H_3}{2H_1}  \hspace{-0.5mm}+ \hspace{-0.5mm} \sqrt{ \frac{1}{4} \left(
	\frac{2H_2^3}{27H_1^3} \hspace{-0.5mm} + \hspace{-0.5mm} \frac{H_3}{H_1} 
    \right)^2 \hspace{-2.5mm} + \hspace{-0.5mm} \frac{1}{27} \left(
    \frac{H_2^2}{3H_1^2}
    \right)^3
  } } \ka
   &\quad +	 \hspace{-0.7mm} \sqrt[3]{ \hspace{-0.5mm} -\frac{H_2^3}{27H_1^3}  \hspace{-0.5mm} - \hspace{-0.5mm} \frac{H_3}{2H_1} \hspace{-0.5mm} -  \hspace{-0.5mm} \sqrt{ \hspace{-0.5mm} \frac{1}{4}\hspace{-0.5mm}  \left(
	\frac{2H_2^3}{27H_1^3} \hspace{-0.5mm} + \hspace{-0.5mm} \frac{H_3}{H_1} 
	\right)^2 \hspace{-2.5mm} +  \hspace{-0.5mm} \frac{1}{27} \hspace{-0.5mm} \left(  \hspace{-0.5mm}
	\frac{H_2^2}{3H_1^2} \hspace{-0.5mm}
	\right)^3
	} } \ka
	& \quad - \frac{H_2}{3H_1}  \label{optimal_I}
	\end{align}
\end{lemma} \
From Lemma \ref{lem:E_I}, we can see that the value of $\SGDrun'$ decreases due to the increased cost of local training $H_1$ as we allocate a larger $\precision$. Since the quantization error decreases as $\precision$ increases, a large $\SGDrun'$ is not required. Hence, an \ac{FL} system can decrease the value of $\SGDrun'$ to reduce the increased local computation energy. We can also see that $\SGDrun'$ increases as the cost of transmission $H_3$ increases. Then, for convergence, the FL algorithm can perform more local iterations instead of frequently exchanging model parameters due to the increased communication overhead. 


Although $\objone$ is non-convex with respect to $\tprecision$, there exists $\tprecision' \in \mathcal{C}$ such that for $\tprecision \leq \tprecision'$, $\objone$ is non-increasing, and for $\tprecision \geq \tprecision'$, $\objone$ is non-decreasing. This is because $\objone$ decreases as the convergence rate becomes faster for increasing $\tprecision$. Then, $\objone$ increases after $\tprecision'$ due to unnecessarily allocated bits. 
Since $\objone$ is differentiable at $\tprecision$, we can find such local optimal $\tprecision'$ from $\partial \objone / \partial \tprecision = 0$ using Fermat's Theorem \cite{HB:16}. To obtain $\tprecision'$, we formulate the transcendental equation as below
\begin{align}
	\frac{\partial \objone}{\partial m} &= 0 \leftrightarrow \tprecision = M_A 2^{2\tprecision} + M_B,
\end{align}
where 
\begin{align}
	M_A &= \frac{\schedulesize(\sum_{k=1}^{\Whole} \prob^2 \sgdvariance_k^2 + 4(\SGDrun-1)^2\sgdbound^2 + 4\SGDrun^2\sgdbound^2/\schedulesize + 4\smooth\Noniid}{4\paramsize\SGDrun\sgdbound^2 \log4} \ka
	&\quad  - \frac{\gamma \frac{\SGDrun(\beta \mu -1)(2\accuracy/\smooth - \beta d (\rho - \mu))/(2^{2\precision}(\beta\mu -1))) )}{\beta^2I}} {4\paramsize\SGDrun\sgdbound^2 \log4}, \ka
	M_B &= \frac{\frac{4 d \SGDrun \sgdbound^2 M_C}{\schedulesize} - 4 \SGDrun^2 \sgdbound^2\log4 \sum_{k=1}^{\Whole} \prob \localE }{4\paramsize\SGDrun\sgdbound^2\log4 M_C} \ka
	M_C &= \sum_{k=1}^{\Whole} \prob \frac{\power_k}{B \log_2\left( 1 +\frac{ \power_k \bar{h}_k}{N_0 B}  \right)}.
\end{align}
We present a closed-form solution of the above equation in the following Lemma.
\begin{lemma}
	For given $\SGDrun$ and $\precision$, the local optimal $\tprecision'$ to minimize $\objone$ will be:
	\begin{align}
		\tprecision' &= M_B - \frac{1}{\log4} W
		\left(
		-M_A \log4 \exp(M_B \log4)
		\right), \label{optimal_m}
	\end{align}

\end{lemma}
where $W(\cdot)$ is the Lambert $W$ function. 

Following the same logic of obtaining $\tprecision'$, we can find a local optimal solution $\precision'$ from the first derivative test. Although there is no analytical solution for $\precision'$, we can still obtain it numerically using a line search method. Then, problem \eqref{obj1} can be optimized iteratively. We first obtain two analytical solutions for $\SGDrun$ and $\tprecision$. From these solutions, we numerically find a local optimal $\precision'$. Since $\objone$ has a unique solution to each variable, it converges to a stationary point \cite{Nonlinearprogramming}. Although these points cannot guarantee to obtain globally Pareto optimal, using the \ac{NBI} method, we are still guaranteed to reach locally Pareto optimal points \cite{NBI}. In Section \ref{sec:simlulation results}, we will also numerically show that the obtained points can still cover most of the practical portion of a global Pareto boundary. For ease of exposition, hereinafter, we refer to these local Pareto optimal points as \textquotedblleft Pareto optimal\textquotedblright. 

\subsection{Minimizing $\objtwo$} \label{sec:minimizing g_2}
Now, we obtain $\bo{x}_2^*$ from the following problem to complete finding the utopia point. 
\begin{subequations}
	\begin{align}
		\min_{\SGDrun, \schedulesize, \tprecision, \precision} \quad \objtwo
		\label{obj2}, \quad
		\textrm{s.t.} \quad (\SGDrun, \schedulesize, \tprecision, \precision) \in \mathcal{C}.
	\end{align}
\end{subequations}
From \eqref{Time_formula}, the objective function is a decreasing function with respect to $\schedulesize, \tprecision$, and $\precision$. Hence, $\Whole, \tpremax$, and $\premax$ are always the optimal solutions to the above problem. Then, the problem can be reduced to a single variable optimization problem with respect to $\SGDrun$. We check the convexity of the reduced problem as follows:
\begin{align}
	\frac{\partial^2 \objtwo}{\partial \SGDrun^2} &= \frac{ \beta^2}{(\beta\strong-1) (\frac{2\accuracy}{\smooth} - \frac{\beta \longconst}{\beta\strong -1} )}  \ka
	&\quad \times
	\left\{
	\sum_{k=1}^{\Whole} \frac{2\prob^2\sgdvariance_k^2}{\SGDrun^3} + \frac{8\sgdbound^2}{\SGDrun^3} + \frac{8\smooth\Noniid}{\SGDrun^3}
	\right\} 
	-\frac{2\gamma}{\SGDrun^3}.
\end{align}
Hence, it is a convex problem for $\gamma < \frac{ \beta^2}{(\beta\strong-1) (\frac{2\accuracy}{\smooth} - \frac{\beta \longconst}{\beta\strong -1} )} 
\left\{
\sum_{k=1}^{\Whole} \prob^2\sgdvariance_k^2 + 4\sgdbound^2 + 4\smooth\Noniid
\right\}$. Since $\gamma$ is an arbitrary constant such that $\gamma \geq 0$, we can always find $\gamma$ that satisfies the above condition. We present a closed-form solution of $\SGDrun$ from the first derivative test in the following lemma.
\begin{lemma} \label{lem:T_I} 
	For $\gamma < \frac{ \beta^2 	\left\{
		\sum_{k=1}^{\Whole} \prob^2\sgdvariance_k^2 + 4\sgdbound^2 + 4\smooth\Noniid
		\right\} }{(\beta\strong-1) (\frac{2\accuracy}{\smooth} - \frac{\beta \longconst}{\beta\strong -1} )} $, the optimal value of $\SGDrun''$ to minimize $\objtwo$ is given by  
	\begin{align}
		\SGDrun'' \hspace{-0.7mm}  = \hspace{-0.7mm} \sqrt{ \hspace{-0.5mm}
			\frac{ \sum_{k=1}^{\Whole} \hspace{-0.5mm} \prob^2 \sgdvariance_k^2 \hspace{-0.5mm} +\hspace{-0.5mm} 4\sgdbound^2 \hspace{-0.5mm} + \hspace{-0.5mm} 4\smooth\Noniid \hspace{-0.5mm} - \hspace{-0.5mm}
			\gamma (\beta\strong-1)
			(
			\frac{2\accuracy}{\smooth} - \frac{\beta \longconst}{\beta\strong -1}
			)
			/\beta^2 }
		{4\sgdbound^2 + \frac{4\sgdbound^2}{\schedulesize}.
		}
	}
	\end{align}
\end{lemma} 
From Lemma \ref{lem:T_I}, we can see that the optimal value of $\SGDrun''$ increases as $\precision$ decreases. This is because the system has to reduce quantization error by training more number of times. 

\subsection{Normal Boundary Inspection} 
We now obtain the Pareto boundary using \ac{NBI}. We redefine $\boldsymbol{\obj}(\SGDrun, \schedulesize, \tprecision, \precision) := \boldsymbol{\obj}(\SGDrun, \schedulesize, \tprecision, \precision) -\boldsymbol{\obj^*}$ so that the utopia point can be located at the origin. The \ac{NBI} method aims to find intersection points between the boundary of $\boldsymbol{\obj}(\SGDrun, \schedulesize, \tprecision, \precision)$ and a normal vector $\hat{\bo{n}} = -\boldsymbol{\payoff} \boldsymbol{1}$, where $\boldsymbol{1}$ denotes the column vector consisting of only ones which are pointing toward the origin. Then, the set of points on such a normal vector will be: $\bo{\payoff} \zeta + s\hat{\bo{n}}$, where $s \in \mathbb{R}$. The intersection points can be obtained from the following subproblem:
\begin{subequations}
	\begin{align}
		\max_{\SGDrun, \schedulesize, \tprecision, \precision, s}  & \quad s
		\label{NBI} \\
		\textrm{s.t.} \quad & \quad (\SGDrun, \schedulesize, \tprecision, \precision) \in \mathcal{C} \\
		& \quad \bo{\payoff} \boldsymbol{\zeta} + s\hat{\bo{n}} = \boldsymbol{\obj}(\SGDrun, \schedulesize, \tprecision, \precision) \label{NBI_feasibility},
	\end{align}
\end{subequations}
where \eqref{NBI_feasibility} makes the set of points on $\bo{\payoff} \zeta + s\hat{\bo{n}}$ be in the feasible area. From the definitions of $\bo{\payoff}$ and $\hat{\bo{n}}$,  constraint \eqref{NBI_feasibility} can be given as 
\begin{align}
	\bo{\payoff} \boldsymbol{\zeta} + s\hat{\bo{n}} &= \begin{bmatrix}
		\obj_1(\bo{x}_2^*) (\zeta_2 - s) \\
		\obj_2(\bo{x}_1^*) (\zeta_1 - s)
		\end{bmatrix} = \begin{bmatrix}
		\objone \\
		\objtwo
	\end{bmatrix} \label{NBI_feasibility_change}.
\end{align}
From \eqref{NBI_feasibility_change}, we obtain the expression of $s$ as below
\begin{align}
	s = \zeta_1 - \frac{\objtwo}{\obj_2(\bo{x}_1^*)} = \zeta_2 - \frac{\objone}{\obj_1(\bo{x}_2^*)}. \label{s_substitute}
\end{align}
Hence, we can change problem \eqref{NBI} as follows
\begin{subequations}
	\begin{align}
		\min_{\SGDrun, \schedulesize, \tprecision, \precision}  & \quad  \frac{\objtwo}{\obj_2(\bo{x}_1^*)} - \zeta_1
		\label{NBI_changed_problem} \\
		\textrm{s.t.} \quad & \quad (\SGDrun, \schedulesize, \tprecision, \precision) \in \mathcal{C} \\
		& \quad 1 - 2\zeta_1 + \frac{\objtwo}{\obj_2(\bo{x}_1^*)} - \frac{\objone}{\obj_1(\bo{x}_2^*)} = 0 \label{NBI_feasibility_changed_changed},
	\end{align}
\end{subequations}
where we substituted $s$ with \eqref{s_substitute} for the objective function, constraint  \eqref{NBI_feasibility_changed_changed} is from \eqref{s_substitute}, and $\zeta_1 + \zeta_2 = 1$. To remove the equality constraint \eqref{NBI_feasibility_changed_changed}, we approximate the problem by introducing a quadratic penalty term $\lambda$ as below 
\begin{subequations}
	\begin{align}
		\min_{\SGDrun, \schedulesize, \tprecision, \precision}  & \quad  \frac{\objtwo}{\obj_2(\bo{x}_1^*)} - \zeta_1 +  \penaltyterm \hspace{-0.5mm} \left( 
		1  -   2\zeta_1  + \frac{\objtwo}{\obj_2(\bo{x}_1^*)} 		\right.  \ka
		&\quad \quad \left. - \frac{\objone}{\obj_1(\bo{x}_2^*)}
		\right)^2
		\label{NBI_changed_changed_problem} \\
		\textrm{s.t.} \quad & \quad (\SGDrun, \schedulesize, \tprecision, \precision) \in \mathcal{C}. 
	\end{align}
\end{subequations}
For $\penaltyterm$, we consider an increasing sequence $\{\penaltyterm_i\}$ with $\penaltyterm_i \rightarrow \infty$ as $i \rightarrow \infty$ to penalize the constraint violation more strongly. We then obtain the corresponding solution $\bo{x}^i$, which is $(\SGDrun, \schedulesize, \tprecision, \precision)$ for minimizing problem \eqref{NBI_changed_changed_problem} with penalty parameter $\penaltyterm_i$. 
\begin{theorem} \label{theorem 2}
	For $\penaltyterm_i \rightarrow \infty$ as $i \rightarrow \infty$, solution $\bo{x}^i$ approaches the global optimal solution of problem \eqref{NBI_changed_changed_problem}, and it also becomes Pareto optimal. 
\end{theorem}
\begin{proof}
	For notational simplicity, we use $\bo{x}$ to denote $(\SGDrun, \schedulesize, \tprecision, \precision) \in \mathcal{C}$.  Let $q^p (\bo{x})$ denote the quadratic penalty term in problem \eqref{NBI_changed_changed_problem}.  We also define a global optimal solution to the problem \eqref{NBI_changed_problem} as $\bar{\bo{x}}$. Since $\bo{x}^i$ minimizes the above problem with penalty parameter $\lambda_i$, we have 
	\begin{align}
	\frac{\obj_2 (\bo{x}^i)}{\obj_2(\bo{x}_1^*)} -\zeta_1 + \lambda_i q^p(\bo{x}^i) &\leq \frac{\obj_2 (\bar{\bo{x}}) }{\obj_2(\bo{x}_1^*)} -\zeta_1 + \lambda_i q^p(  \bar{\bo{x}}) \ka
	&\leq \frac{\obj_2 (\bar{\bo{x}}) }{\obj_2(\bo{x}_1^*)} -\zeta_1,
	\end{align}
where the last inequality is from the fact that $\bar{\bo{x}}$ minimizes problem \eqref{NBI_changed_problem} with the equality constraint of $q^p(\bo{\bar{x}})$ being zero. Then, we obtain the inequality of $q^p(\bo{x}^i)$ as follows
\begin{align}
	q^p(\bo{x}^i) \leq \frac{1}{\lambda_i} \left(
	\frac{\obj_2 (\bar{\bo{x}}) }{\obj_2(\bo{x}_1^*)} - \frac{\obj_2 (\bo{x}^i)}{\obj_2(\bo{x}_1^*)}
	\right).
\end{align}
By taking the limit as $i \rightarrow \infty$, we have
\begin{align}
	\limit{i} q^p(\bo{x}^i) \leq \limit{i} \frac{1}{\lambda_i} \left(
	\frac{\obj_2 (\bar{\bo{x}}) }{\obj_2(\bo{x}_1^*)} - \frac{\obj_2 (\bo{x}^i)}{\obj_2(\bo{x}_1^*)}
	\right) = 0.
\end{align}
Hence, as $\penaltyterm_i \rightarrow \infty$, we can see that $\bo{x}^i$ approaches the global optimal solution of \eqref{NBI_changed_problem}, which aims to find a Pareto optimal point. 
\end{proof}
From Theorem 2, we can obtain a global optimal solution of (42a), and this correspond to a Pareto optimal point for specific values of $\zeta_1$ and $\zeta_2$. Note that problem (43a) can be solved using a software solver. To fully visualize the boundary, we iterate problem (39a) for various combinations of $\zeta_1$ and $\zeta_2$. The overall algorithm is given in Algorithm 2.
	
The main complexity of Algorithm 2 at each iteration is to solve problem (42a), which corresponds to line $9-13$. We approximated problem (42a) to problem (43a), which can be solved by a software solver. If we use the interior point method with a desired accuracy $\accuracy_\text{in}$, then the complexity can given by $\mathcal{O}(\log(\frac{1}{\accuracy_\text{in}}))$ \cite{Convex}. Since we solve (43a) by increasing $\penaltyterm_i$ at iteration $i$, the complexity of this outer loop can be given by $\mathcal{O}(\log(\frac{1}{\accuracy_\text{out}}))$ with a desired accuracy $\accuracy_\text{out}$. Therefore, the complexity of Algorithm 2 is $\mathcal{O}(\log(\frac{1}{\accuracy_\text{out}}) \log(\frac{1}{\accuracy_\text{in}} ))$.
\begin{algorithm}[t!] 
	\caption{\ac{NBI} approach to obtain Pareto boundary}
	\KwInput{$\Whole, B, \power, \SGDrun_{\text{min}}, \schedulesize_{\text{min}}, \tpremax, \premax, \beta, \gamma, \sgdbound, \sgdvariance, \strong, \smooth, A, \alpha$, accuracy constraint $\accuracy$, loss function $\loss (\cdot)$, stopping criterion $\accuracy_{\text{uto}}$ and $\accuracy_\text{out}$, and a structure of \ac{QNN}}
	
	To find $\bo{\obj}_1^*$, initialize $(\SGDrun, \schedulesize, \tprecision, \precision)$ and set $\schedulesize = \Whole$\\
	\While{$\sqrt{(\SGDrun - \SGDrun')^2 + (\tprecision - \tprecision')^2 + (\precision - \precision')^2} > \accuracy_{\text{uto}}$ } {
		Update $(\SGDrun, \tprecision, \precision)$ as $(\SGDrun', \tprecision', \precision')$ \\
		Obtain $\SGDrun$' from \eqref{optimal_I} \\
		Obtain $\tprecision'$ for fixed $\SGDrun'$ from \eqref{optimal_m} \\
		Obtain $\precision'$ for fixed $\SGDrun'$ and $\tprecision'$ using a line search		
	} 
	To find $\bo{\obj}_2^*$, calculate $\SGDrun''$ from Lemma \ref{lem:T_I} and set $(\schedulesize, \tprecision, \precision) = (\Whole, \tpremax, \premax)$ \\
\While{$\zeta_1 \leq 1$}{
	Initialize $\bo{x}$, which denotes a vector $(\SGDrun, \schedulesize, \tprecision, \precision)$
\Repeat{$\sqrt{||\bo{x} - \bo{x}'||^2} \leq \accuracy_{\text{out}}$ }{
	Update $\bo{x}$ as  $\bo{x}'$\\
	Obtain $\bo{x}'$ from problem \eqref{NBI_changed_changed_problem} \\
    Increase $\lambda$
}
	Round $(\SGDrun, \schedulesize, \tprecision, \precision)$ and increase $\zeta_1$
}
\end{algorithm} 
\vspace{-0.5cm}

\subsection{Nash Bargaining Solution} \label{sec:NBS}
Since the solutions from \eqref{obj} are Pareto optimal, there is always an issue of choosing the best point. This is because any improvement on one objective function leads to the degradation of another. We can tackle this problem considering a bargaining process \cite{gamethoerybook} between two players: one tries to minimize the energy consumption and another aims to reduce the number of communication rounds. Since the parameters of \ac{FL}, i.e.,  $(\SGDrun, \schedulesize, \tprecision, \precision)$, are shared, the players should reach a certain agreement over the parameters.
It is known that \ac{NBS} can be a unique solution to this bargaining process. The \ac{NBS} was chosen here because it satisfies several fairness axioms \cite{gamethoerybook}, and thus, it has been used as a fair solution to resource management problems \cite{LE:08, PH:07}. We can obtain the \ac{NBS} from the following problem \cite{gamethoerybook}:
\begin{subequations}
	\begin{align}
		\max_{\obj_1 (\bo{x}), \obj_2(\bo{x})}  & \quad (\obj_1(\disagreement) - \obj_1(\bo{x}))(\obj_2(\disagreement) - \obj_2(\bo{x}) )
		\label{NBS} \ka
	   \textrm{s.t.}  \quad  & \quad (\obj_1(\bo{x}), \obj_2(\bo{x})) \in \overline{\achievable},
	\end{align}
\end{subequations}
where $\achievable = \underset{\bo{x} \in \mathcal{C}}{\cup} \ (\obj_1 (\bo{x}), \obj_2(\bo{x}))$ is the achievable set of $(\obj_1(\bo{x}), \obj_2(\bo{x}))$, $\overline{\achievable}$ represents the convex hull of $\achievable$, and $\disagreement$ is the outcome when the players fail to cooperate. Since the \ac{NBS} always lies on the Pareto boundary, we perform the bargaining process on the obtained boundary from Algorithm 2. Then, we can find the \ac{NBS} graphically by finding a tangential point where the boundary and a parabola $(\obj_1(\disagreement) - \obj_1(\bo{x}))(\obj_2(\disagreement) - \obj_2(\bo{x}) = \Delta$ intersects with constant $\Delta$. 

\section{Simulation Results and Analysis} \label{sec:simlulation results}
	For our simulations, unless stated otherwise, we uniformly deploy $\Whole = 50$ devices over a square area of size $500$ m $\times$ $500$ m serviced by one \ac{BS} at the center, and we assume a Rayleigh fading channel with a path loss exponent of 4. We assume that the \ac{FL} algorithm is used for a classification task with MNIST dataset. We distribute the training dataset over devices in a non-iid fashion by allocating labels from a Dirichlet distribution with parameter 0.1.  A softmax classifier is used to measure our \ac{FL} performance. We also use $\power_k = 100$ mW, $B = 10$ MHz, $N_0 = -173$ dBm, $S= 2$ MB, $x_\text{in} = 786$,  $\tpremax =32$ bits, $\premax =32$ bits, $\SGDrun_{\text{min}} = 1$, $\SGDrun_{\text{max}} = 30$, $\schedulesize_{\text{min}} = 1$, $\accuracy = 0.1$, $\smart = 100$, and $\gamma = 1$, $\forall k = 1, \dots, N$. For $\smooth$, we used the reported value $\smooth = 0.097$ with the same dataset and the loss function \cite{YE:21}. However, estimating $\strong$ is more challenging than the estimation of $\smooth$. Since the value of $\strong$ is widely assumed to be a small value between $[0.001, 1]$ \cite{HO:15}\cite{convexity}, we used $\strong = 0.05$ as done in \cite{convexity} with the same dataset. We assume that each device trains a \ac{QNN} structure with five convolutional layers and three fully-connected layers. Specifically, the convolutional layers consist of  128 kernels of size 3 × 3, two of 64 kernels of size 3×3, and two of 32 kernels of size 3x3. The first layer is followed by 3x3 pooling and the second and the fifth layer are followed by 3x3 max pooling with a stride of two. Then, we have
	one dense layer of 2000 neurons, one fully-connected layer of 100 neurons, and the output layer. In this setting, we have $N_c = 0.0405 \times 10^9, \paramsize = 0.41 \times 10^6$, and $O_s = 4990.$ To estimate $\sgdbound$ and $\sgdvariance_k$, we measure every device's average maximum norm of stochastic gradients $\sgdbound_k$ for the initial 20 local iterations and set $\sgdbound = \max_{k} \ \sgdbound_k$, $\forall k = \{1, \dots, \Whole\}.$ We used the same gradients information to estimate $\sgdvariance_k$ while measuring $G_k$. Since the norm of the stochastic gradient generally decreases with training epochs, we use the initial values of $\sgdbound_k$ to estimate $\sgdbound$ as in \cite{OL:15}. Similarly, since loss functions are in general decreasing with training epoch, we can bound $\Noniid$ as $\Noniid =  \sum_{k=1}^{\Whole} \prob (\loss(\weights^*)  - \loss^*) \leq \sum_{k=1}^{\Whole} \prob \loss(\weights^*) \leq \sum_{k=1}^{\Whole} \prob \loss(\weights')$, where $\weights'$ can be a global model in early stage. From the above setting, we estimated $\sgdbound = 0.25$. We then used the global model, which was used to measure $\sgdbound$, to estimate $\Noniid = 0.6$. For the computing model, we use a $28$ nm technology processing chip and set $\MACamplitude = 3.7$ pJ, $\DRAMCO = 150$, and $\MACex = 1.25$ as done in \cite{MB:18}. For the disagreement point $\disagreement$, we use $(\SGDrun_{\text{max}}, 1, 1, \precision_\text{min})$ as this setting is neither biased towards minimizing the energy consumption nor towards the number of communication rounds. We assume that each device has the same architecture of the processing chip. All statistical results are averaged over a number of  independent runs. 

\begin{figure}
	\centering
	\begin{center}
		\psfrag{xxxxxxxxxxxxxxx1}[bc][bc][0.5]{Energy consumption $[J]$}
		\psfrag{yyyyyyyyyyyyyyyyyy2}[tc][tc][0.5]{The number of communication rounds}
		\psfrag{a1111111111111111}[Bl][Bl][0.59]{Exaustive search}
		\psfrag{b22222222222}[Bl][Bl][0.59]{Proposed algorithm}
		\includegraphics[width=1.00\columnwidth]{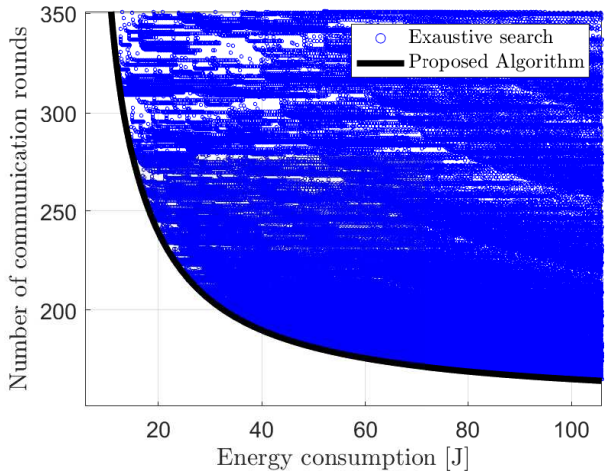}
	\end{center}
\vspace{-0.0cm}
\caption{Pareto Boundary from Algorithm 2 and feasible area from exhaustive search}
\vspace{-0.0cm}
\label{fig:exaustive_search}
\end{figure}

Figure \ref{fig:exaustive_search} shows the Pareto boundary from Algorithm 2 as well as the feasible area obtained from the exhaustive search for $\Whole = 50$. We can see that our boundary and the actual Pareto boundary match well. Although we cannot find the global Pareto optimal points due to the non-convexity of problem  \eqref{obj1}, it is clear that our analysis can still cover most of the important points that can effectively show the tradeoff in the feasible region. 

\begin{figure*}[t!]
	\begin{subfigure}[t]{0.33\textwidth}
			\begin{center}   
			{ 
				\psfrag{A11111}[Bl][Bl][0.59]   {NBS}
				\psfrag{A2}[Bl][Bl][0.59]   {SUM}
				\psfrag{X111111111111111111111}[bc][bc][0.5] {Energy consumption $[J]$}
				\psfrag{Y11111111111111111111111111111}[tc][tc][0.5] {The number of communication rounds}
				\includegraphics[width=\textwidth]{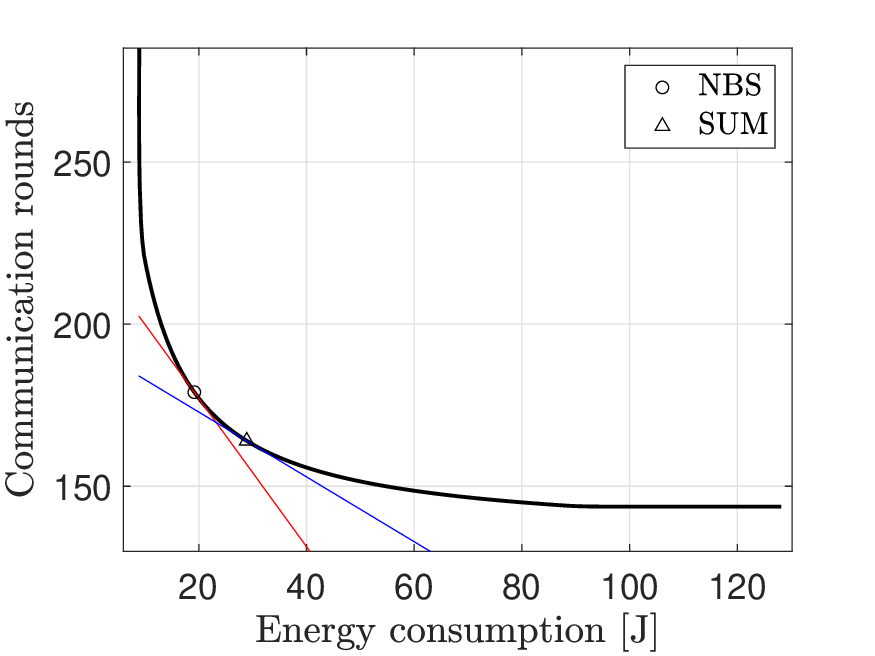}
			}
		\end{center}
		\vspace{-0.0cm}
		\caption{$\Whole$ = 10}
		\label{fig:N=10}
	\end{subfigure}\hfill
	\begin{subfigure}[t]{0.33\textwidth}
			\begin{center}   
	{ 
		\psfrag{A11111}[Bl][Bl][0.59]   {NBS}
		\psfrag{A2}[Bl][Bl][0.59]   {SUM}
		\psfrag{X111111111111111111111}[bc][bc][0.5] {Energy consumption $[J]$}
		\psfrag{Y11111111111111111111111111111}[tc][tc][0.5] {The number of communication rounds}
		\includegraphics[width=\textwidth]{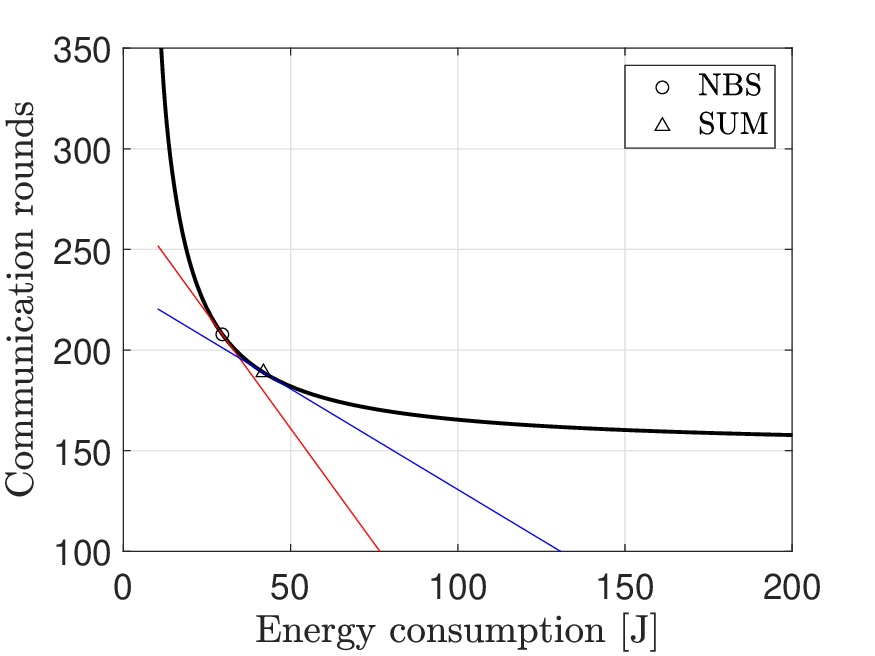}
	}
		\end{center}
		\vspace{-0.0cm}
		\caption{$\Whole$ = 50}
		\label{fig:N=50}
	\end{subfigure}\hfill
	\begin{subfigure}[t]{0.33\textwidth}
			\begin{center}   
	{ 
		\psfrag{A11111}[Bl][Bl][0.59]   {NBS}
		\psfrag{A2}[Bl][Bl][0.59]   {SUM}
		\psfrag{X111111111111111111111}[bc][bc][0.5] {Energy consumption $[J]$}
		\psfrag{Y11111111111111111111111111111}[tc][tc][0.5] {The number of communication rounds}
		\includegraphics[width=\textwidth]{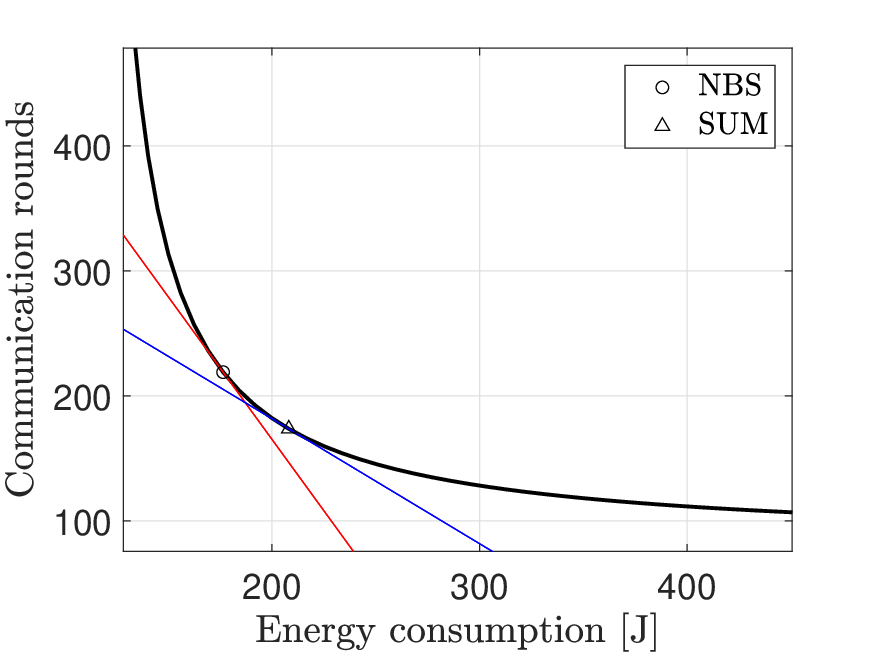}
	}
			\end{center}
		\vspace{-0.0cm}
		\caption{$\Whole$ = 200}
		\label{fig:N=200}
	\end{subfigure}\hfill
	\vspace{-0.0cm}
	\caption{Pareto boundaries, \ac{NBS}, and SUM points for varying the number of devices $\Whole$.}
	\label{fig:Pareto boundary}
	\vspace{-0.0cm}
\end{figure*}

\begin{table}
\begin{subtable}[c]{0.5\textwidth}
	\centering
		\scalebox{1.0}{
		\begin{tabular}{|c|c|c|c|c|}
		\hline
		& $\Whole = 10$     & $\Whole = 50$     & $\Whole = 200$     \\ \hline
		\ac{NBS}      & $(2, 2, 12, 19)$  & $(1, 5, 12, 19)$   & $(1, 23, 12, 19)$  \\ \hline
		SUM      & $(3, 3, 12, 20)$  & $(1, 7, 12, 20)$  & $(1, 35. 12. 19)$  \\ \hline
		$E_\text{min}$ & $(1, 1, 10, 15)$  & $(1, 1, 11, 15)$  & $(1, 1, 12, 15)$   \\ \hline
		$T_\text{min}$ & $(3, 10, 32, 32)$ & $(1, 50, 32, 32)$  & $(1, 200, 32, 32)$ \\ \hline 
	\end{tabular}}
	\subcaption{Solutions for varying $\Whole$}
	\label{table:N} \vspace{-0.0cm}
\end{subtable}
\begin{subtable}[c]{0.5\textwidth}
	\centering
		\scalebox{1.0}{
		\begin{tabular}{|c|c|c|c|c|}
		\hline
		& CNN1     & CNN2   & CNN3     \\ \hline
		\ac{NBS}      & $(1, 3, 11, 19)$  & $(1, 5, 12, 19)$   & $(1, 8, 14, 20)$  \\ \hline
		SUM      & $(1, 4, 11, 20)$  & $(1,7, 12, 20)$   & $(1, 2, 14, 20)$  \\ \hline
		$E_\text{min}$ & $(1, 1, 10, 15)$  & $(1, 1, 11, 15)$    & $(1, 1, 14, 16)$   \\ \hline
		$T_\text{min}$ & $(2, 50, 32, 32)$ & $(1, 50, 32, 32)$ & $(1, 50, 32, 32)$ \\ \hline
	\end{tabular}}
	\subcaption{Solutions for varying model size}
	\label{table:CNN} \vspace{-0.0cm}
\end{subtable}
\caption{Solutions of \ac{NBS}, SUM, $E_\text{min}$, and $T_\text{min}$ for varying $\Whole$ and the model size.}
\label{table:solution} \vspace{-0.0cm}
\end{table}

Figure \ref{fig:Pareto boundary} and Table \ref{table:N} show the Pareto boundaries obtained from the Algorithm 2 and the solutions of four possible operating points, respectively, for varying $\Whole$. Each solution represents $(\SGDrun, \schedulesize, \tprecision, \precision)$, where $\SGDrun$ is the number of local iterations, $\schedulesize$ is the number of sampled devices, $\tprecision$ is the precision level for transmission, and $\precision$ is the precision level for local training. \ac{SUM} represents the point that minimizes the sum of the two objectives. We can obtain the \ac{SUM} by finding a tangential point between the Pareto boundary and the line $\objone + \objtwo = \Delta$ with $\Delta \in \mathbb{R}$ using a bisection algorithm.  $E_\text{min}$ and $T_\text{min}$ are the solutions that separately optimize $\objone$ and $\objtwo$, respectively. From Fig. \ref{fig:Pareto boundary}, we can see that the energy consumption increases while the number of communication rounds decreases to achieve the target accuracy for increasing $\Whole$. The \ac{FL} system can choose more devices at each communication round as $\Whole$ increases. Hence, the impact of \ac{SG} variance decreases as shown in Theorem \ref{Thm1}. Since involving more devices in the averaging process implies an increase in the size of the batch, the convergence rate increases by using more energy \cite{SA:20}.  

From Table \ref{table:N} and Fig. \ref{fig:Pareto boundary}, we can see that \ac{NBS} points are more biased toward reducing the energy consumption while the \ac{SUM} points focus on minimizing communication rounds. We can also see that, as $\Whole$ becomes larger, the optimal $\SGDrun$ decreases while $\schedulesize$ increases. This is because $\SGDrun$ is a decreasing function with respect to $\sgdbound$ as shown in Lemmas \ref{lem:E_I} and \ref{lem:T_I}. Hence, the \ac{FL} system decreases $\SGDrun$ to avoid model discrepancy over devices since the estimated value of $\sgdbound$ becomes larger for increasing $\Whole$. However, a small $\SGDrun$ will slow down the process to reach optimal weights in the local training. To mitigate this, the \ac{FL} system then increases $\schedulesize$ so that it can obtain more information in the averaging process by selecting more devices.
	
\begin{figure*}[t!]
	\begin{subfigure}[t]{0.33\textwidth}
		\centering	
		\includegraphics[width=\textwidth]{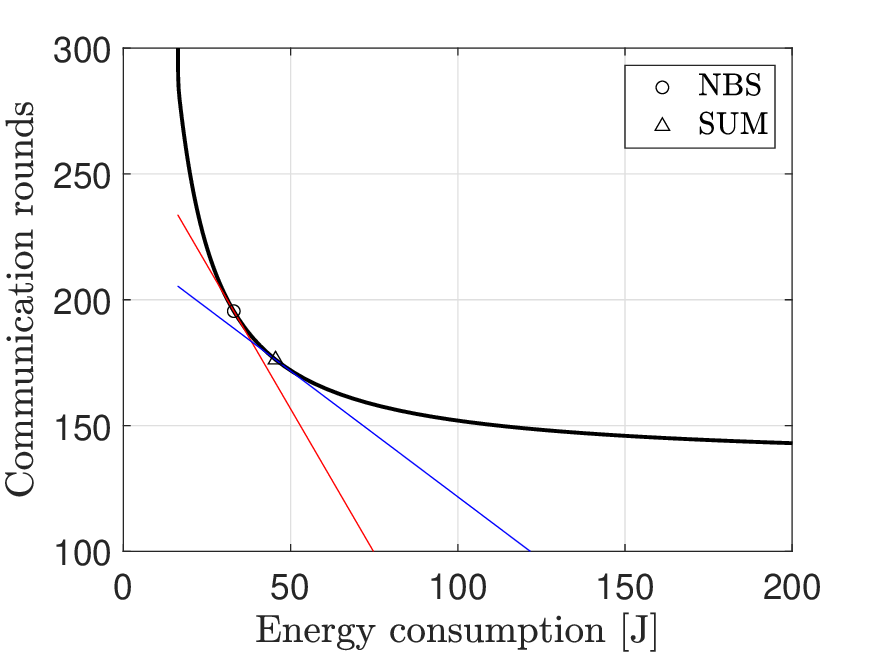}
		\caption{CNN model 1}
		\label{fig:CNN_1}
	\end{subfigure}\hfill
	\begin{subfigure}[t]{0.33\textwidth}
		\centering
		\includegraphics[width=\textwidth]{Figures/CNN_baseline.eps}
		\caption{CNN model 2}
		\label{fig:CNN_2}
	\end{subfigure}\hfill
	\begin{subfigure}[t]{0.33\textwidth}
		\centering
		\includegraphics[width=\textwidth]{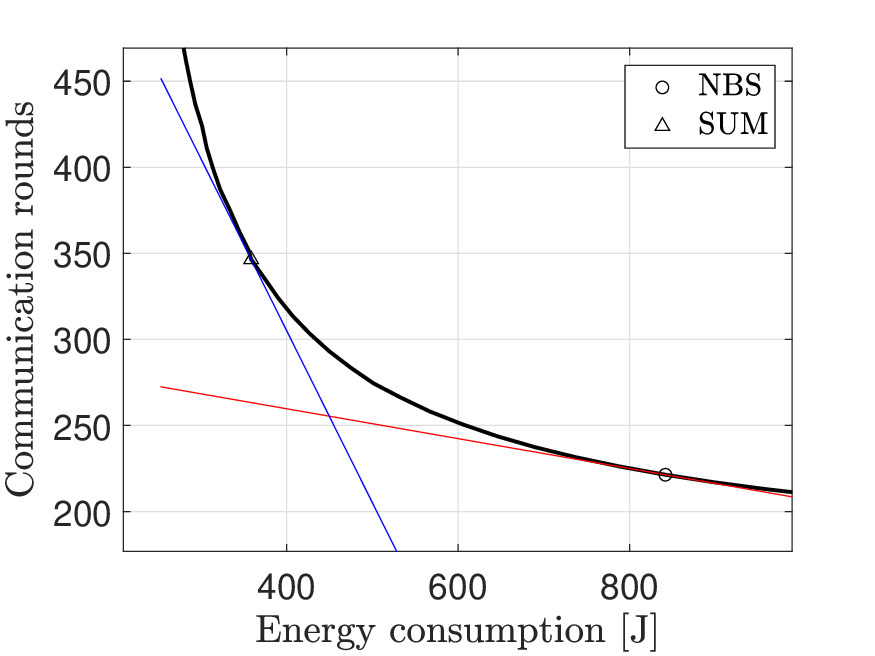}
		\caption{CNN model 3}
		\label{fig:CNN_4}
	\end{subfigure}\hfill
	\vspace{-0.0cm}
	\caption{Pareto boundaries, \ac{NBS}, and SUM points for varying the model size.}
		\vspace{-0.0cm}
	\label{fig:CNN}
\end{figure*}

Figure \ref{fig:CNN} and Table \ref{table:CNN} present the Pareto boundaries from the Algorithm 2 and the corresponding solutions when increasing the size of the neural networks. We keep the same structure of our default \ac{CNN}, but we now increase the number of neurons in the convolutional and fully-connected layers. For each CNN model, the number of parameters will be $0.27\times10^6, 0.41\times10^6,$ and $1.61\times10^6$, respectively. Fig. \ref{fig:CNN} and Table \ref{table:CNN} show that the energy consumption and the number of communication rounds until convergence increase with the model size. For CNN3, the energy cost increased significantly since its large model size cannot be fit into the SRAM even after quantization. From Table \ref{table:CNN}, we can see that the \ac{FL} system requires higher precision levels for larger neural networks. This is because the quantization error increases for larger neural networks, as per Lemma \ref{Lem1}. Hence, the \ac{FL} system allocates more bits for both the computation and the transmission so as to mitigate the quantization error. This, in turn, means that the use of larger neural networks will naturally require more energy, even if the neural network is quantized.

\begin{figure*}[t!]
	\centering	
	\begin{subfigure}[t]{0.43\textwidth}
		\centering	
		\includegraphics[width=\textwidth]{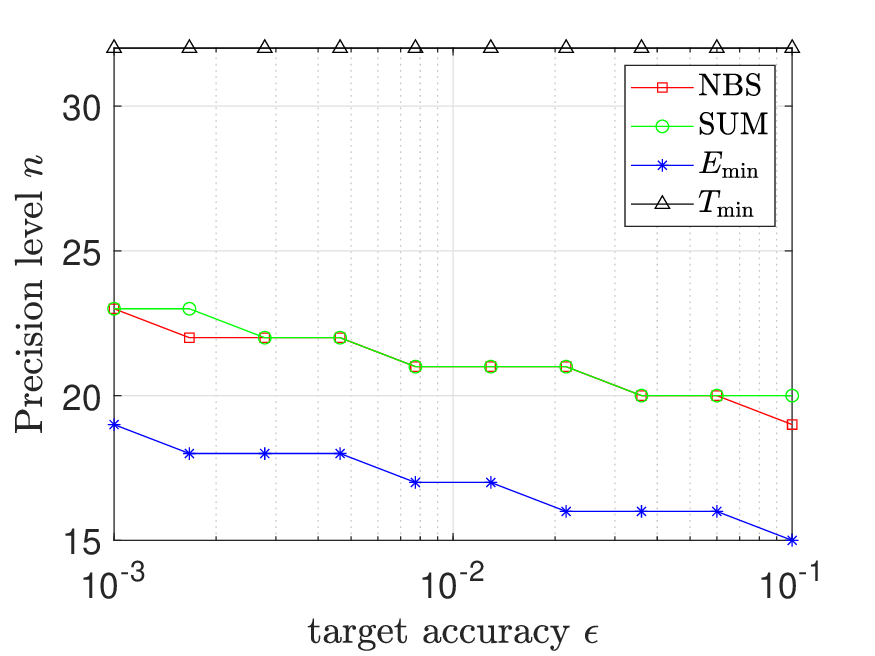}
		\caption{Impact of target accuracy $\accuracy$ on $\precision$}
		\label{fig:epsilon_n}
	\end{subfigure}\hfill
	\begin{subfigure}[t]{0.43\textwidth}
		\centering
		\includegraphics[width=\textwidth]{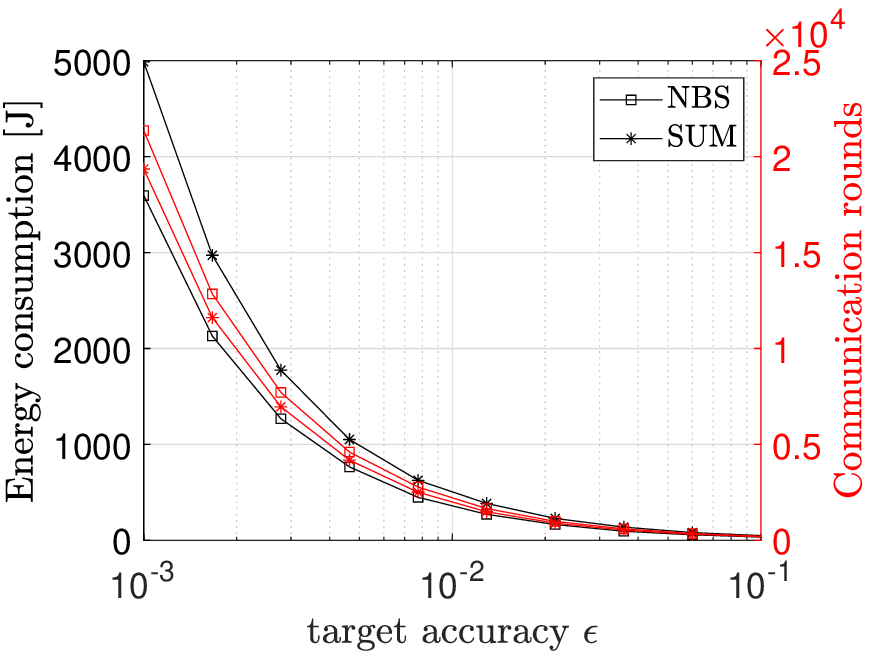}
		\caption{Impact of target accuracy $\accuracy$ on the performance}
		\label{fig:epsilon_performance}
	\end{subfigure}\hfill
	\vspace{-0.0cm}
	\caption{Impact of target accuracy $\accuracy$ on the optimal precision level $\precision$ and the performance}
	\label{fig:epsilon}
	\vspace{-0.0cm}
\end{figure*}

Figure \ref{fig:epsilon_n} presents the impact of target accuracy $\accuracy$ on the optimal precision level $\precision$ and the performance. We can see that as target accuracy $\accuracy$ increases, the optimal precision level $\precision$ also increases to achieve the convergence. This is because quantization in local training yields the unavoidable term $\longconst$ as shown in Theorem \ref{Thm1}. Hence, to achieve a higher target accuracy $\accuracy$, we need to allocate more precision level $\precision$ for local training to achieve the convergence. However, this can increase the number of DRAM accesses to fetch model parameters due to the increased memory size. Since the DRAM access energy is much larger than the \ac{MAC} operation energy, the energy consumption may increase significantly. In Fig. \ref{fig:epsilon_performance}, we can see that we need much more energy and communication rounds to achieve a higher target accuracy $\accuracy$.

\begin{figure*}[t!]
	\centering	
	\begin{subfigure}[t]{0.43\textwidth}
		\centering	
		\includegraphics[width=\textwidth]{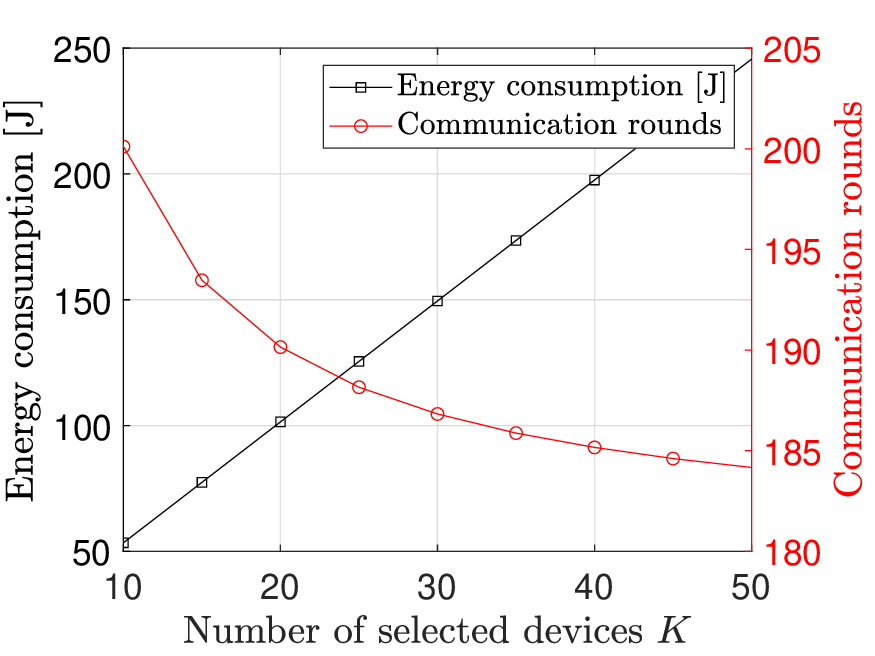}
		\caption{Performance of \ac{NBS} with different $K$}
		\label{fig:K_NBS}
	\end{subfigure} \hfill
	\begin{subfigure}[t]{0.43\textwidth}
		\centering
		\includegraphics[width=\textwidth]{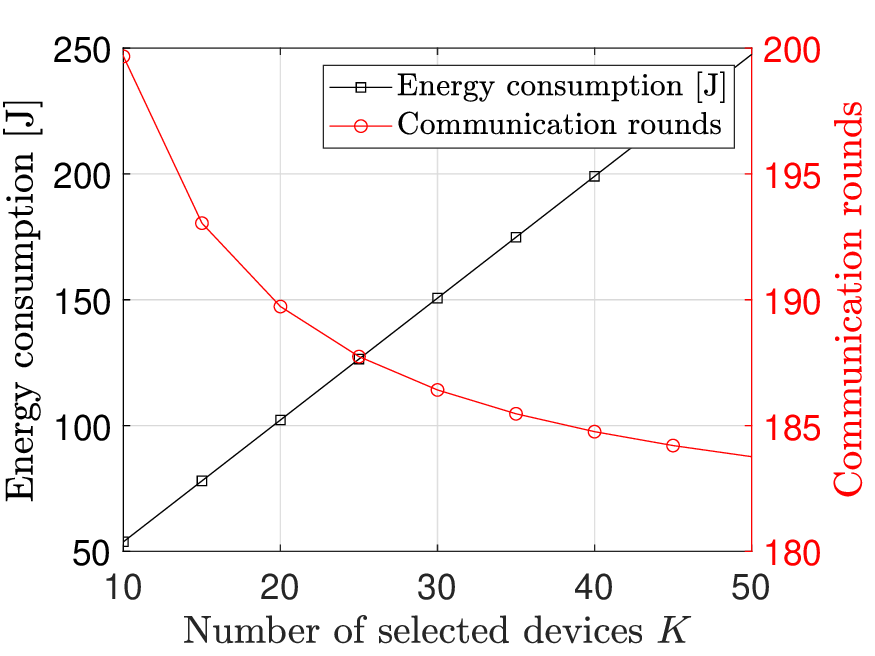}
		\caption{Performance of \ac{SUM} with different $K$}
		\label{fig:K_SUM}
	\end{subfigure}\hfill
	\vspace{-0.0cm}
	\caption{Performance of \ac{NBS} and \ac{SUM} points for increasing $K$.}
	\vspace{-0.0cm}
	\label{fig:change_K_performance}
	\vspace{-0.0cm}
\end{figure*}

In Fig. \ref{fig:change_K_performance}, we show the performance of the \ac{NBS} and the \ac{SUM} points with increasing $\schedulesize$. We can see that the required communication rounds decrease as $\schedulesize$ increases for both schemes. Hence, we can improve the convergence rate by increasing $\schedulesize$ at the expense of more energy. This corroborates the analysis in Section \ref{sec:minimizing g_1}, which shows the total energy consumption is linear with respect to $\schedulesize$. Similarly, it also corroborates the fact that the required number of communication rounds to achieve a certain $\accuracy$ is a decreasing function of $\schedulesize$, i.e., $\mathcal{O}(\frac{1}{\schedulesize})$, in Section \ref{sec:minimizing g_2}. However, we can see that this improvement is not much beneficial \cite{Xi:20} as it linearly increases the energy consumption. 

\begin{figure*}
	\centering	
	\begin{subfigure}[t]{0.43\textwidth}
		\centering	
		\includegraphics[width=\textwidth]{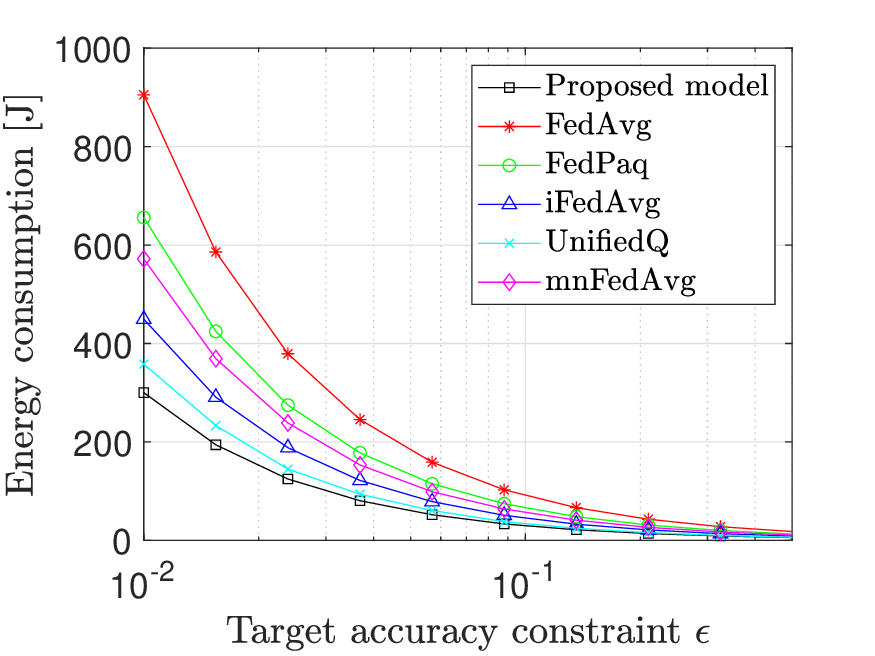}
		\caption{Energy consumption to achieve $\accuracy$}
		\label{fig:Energy}
	\end{subfigure}\hfill
	\begin{subfigure}[t]{0.43\textwidth}
		\centering
		\includegraphics[width=\textwidth]{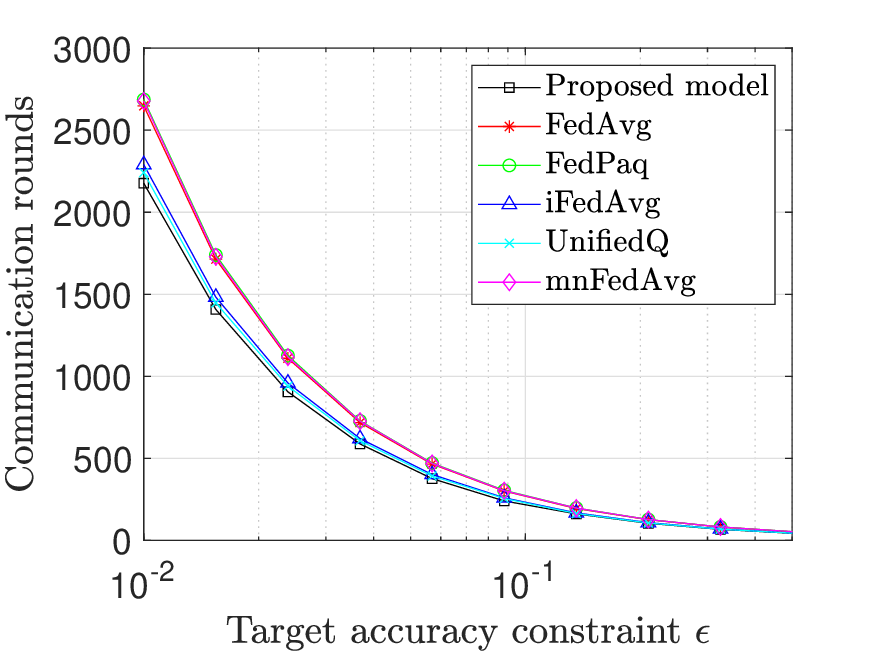}
		\caption{Communication rounds to achieve $\accuracy$ }
		\label{fig:T}
	\end{subfigure}\hfill
	\vspace{-0.0cm}
	\caption{Performance comparison between the proposed model and the baselines to achieve $\accuracy$.}
	\vspace{-0.0cm}
	\label{fig:baselines}
	\vspace{-0.0cm}
\end{figure*}
	Figure 8 shows the required energy and communication rounds to achieve $\epsilon$ using the \ac{NBS} points. For FedAvg \cite{HB:16}, we use $(2, 5, 32, 32)$. FedPaq algorithm \cite{RE:20} uses periodic averaging, partial client participation, and quantization in transmission. Hence, we only optimize $\tprecision$ and use the same setting as FedAvg.  iFedAvg scheme is proposed in \cite{BL:21}, and it optimizes $(\SGDrun, \schedulesize)$ while data is represented in full-precision. UnifiedQ is a baseline introduced in the work in \cite{Ch:22}, and it optimizes $(\SGDrun, \precision)$. We use $\tprecision = 16$ as done in \cite{Ch:22}. Here, we set $K = 5$ for a fair comparison while the original version sampled whole devices at each round. For mnFedAvg, we only optimize $(\tprecision, \precision)$ with $(\SGDrun, \schedulesize) = (2, 5)$. All optimal parameters of the baselines are obtained from solving (18a). From Figs. 8a and 8b, we can see that our algorithm is the most efficient because it consumes the least energy and converges faster than other baselines to achieve $\accuracy$. This is because we optimize all system parameters $(\SGDrun, \schedulesize, \tprecision, \precision)$ simultaneously. From Baseline 1 and 2, we observe that quantization during transmission is beneficial to save the energy, and it does not significantly affect the convergence rate. In particular, we can achieve around 70\% of energy savings compared to FedAvg and around 16\% of energy savings compared to UnifiedQ.

\section{Conclusion} \label{sec:conclusion}
	\vspace{-0.0mm}
	In this paper, we have studied the problem of energy-efficient quantized \ac{FL} over wireless networks. We have presented the energy model for our \ac{FL} based on the physical structure of a processing chip considering the quantization. Then, we have formulated a multi-objective optimization problem to minimize the energy consumption and the number of communication rounds simultaneously under a target accuracy by controlling the number of local iterations, the number of selected users, the precision levels for the transmission, and the training. To solve this problem, we first have derived the convergence rate of our quantized \ac{FL}. Based on it, we have used the \ac{NBI} method to obtain the Pareto boundary. We also have derived analytical solutions that can optimize each objective function separately. Simulation results have validated our theoretical analysis and provided design insights with two practical operating points. We have also shown that our model requires much less energy than a standard \ac{FL} model and the baselines to achieve the convergence. 
	In essence, this work provides the first systematic study on how to optimally design quantized \ac{FL} balancing the tradeoff between energy efficiency and convergence rate, and the target accuracy over wireless networks.
	
	\begin{appendix}
	\subsection{Additional Notations} \label{Additional Notations}
	As done in \cite{Xi:20}, we define $t$ as the round of the local iteration with a slight abuse of notation. Then,  $\weights^k_t$ becomes the model parameter at local iteration $t$ of device $k$. If $t \in \globalset$, where $\globalset = \{j \SGDrun \ | \ j= 1, 2, \dots\}$, each device transmits model update $\qmodelupdate_t$ to the \ac{BS}. We introduce an auxiliary variable $\vweights^k_{t+1}$ to represent the result of one step of local training from $\weights^k_t$. At each local training, device $k$ updates its local model using \ac{SGD} as below
	\begin{align}
		\vweights^k_{t+1} = \weights^k_t - \learningrate_t \nabla \globall_k (\qweights_t, \minibatch^k_t). \label{v_sgd_update}
	\end{align}
	The result of the $(t+1)$th local training will be $\weights^k_{t+1} = \vweights^k_{t+1}$ if $t+1 \not\in \globalset$ because device $k$ does not send a model update to the \ac{BS}. If $t+1 \in \globalset$, each device calculates and transmits its model update, and then the global model is generated as $\weights_{t+1} = \weights_{t - \SGDrun +1} + \frac{1}{K}\sum_{k\in \mathcal{N}_{t+1}} \qmodelupdate_{t+1}$. Note that $\qmodelupdate_{t+1} = Q(\vweights^k_{t+1} - \weights_{t-\SGDrun+1})$ and $\weights_{t - \SGDrun +1}$ is the most recent global model received from the \ac{BS}. We provide the aforementioned cases below:
	\begin{align}
		&\weights^k_{t+1}  = \begin{cases}
			\vweights^k_{t+1} \quad &\text{if} \quad t+1 \not\in \globalset, \\
			\weights_{t-\SGDrun+1} + \frac{1}{K} \sum_{k\in \mathcal{N}_{t+1}}\qmodelupdate_{t+1} \quad &\text{if} \quad t+1 \in \globalset.
		\end{cases}
	\end{align}
    Now, we define two more auxiliary variables:
	%
	%
	$\avgv_t = \sum_{k=1}^{\Whole} \prob \vweights^k_t$ and $\avgweights_t =  \sum_{k=1}^{\Whole} \prob \weights^k_t$. Similarly, we denote $\avgsgd_t = \sum_{k=1}^{\Whole} \prob  \nabla \globall_k(\qweights_t, \minibatch^k_t)$ and $\avggd_t = \sum_{k=1}^{\Whole} \prob \nabla \globall_k(\qweights_t)$.
	From \eqref{v_sgd_update}, we can see that $\avgv_{t+1} = \avgweights_t - \learningrate_t \avgsgd_t$. 
	
	\subsection{The result of one local iteration} \label{proof_lem_2}
	We present a preliminary lemma to prove Theorem 1. We first present the result of one iteration of local training in the following lemma.
	\begin{lemma}
		Under Assumption \ref{assumption 1}, we have 
		\begin{align}
			\E  \hspace{-0.0mm} \left[ \hspace{-0.0mm} 
			||\avgv_{t+1}  \hspace{-0.5mm} - \hspace{-0.5mm} \weights^*||^2
			\right] 
			&\hspace{-0.5mm} \leq \hspace{-0.5mm} (1 \hspace{-0.0mm} - \hspace{-0.0mm}\strong \learningrate_t) \E \hspace{-0.0mm}
			\left[ \hspace{-0.0mm} 
			||\avgweights_t \hspace{-0.0mm} - \hspace{-0.0mm} \weights^*||^2
			\right] \hspace{-0.0mm}
			+ \hspace{-0.0mm} \frac{\learningrate_t d(\smart \hspace{-0.0mm}  - \hspace{-0.0mm}  \strong)}{2^{2\paramsize}} \ka
			&\quad			+ \hspace{-0.5mm}  \learningrate_t^2  
			\left(
			\sum_{k=1}^{\Whole} \prob^2 \sgdvariance_k^2 
			 + 4(\SGDrun  - 1)^2 \sgdbound^2 \hspace{-0.5mm}
			+ \hspace{-0.5mm}  4\smooth \Noniid
			\right) . 
		\end{align}\label{Lem2}
	\end{lemma}
	\begin{proof}
		From $\avgv_{t+1} = \avgweights_t - \learningrate_t \avgsgd_t$, we have 
		\begin{align}
			||\avgv_{t+1} - \weights^*||^2 &= ||\avgweights_t - \learningrate_t \avgsgd_t - \weights^* - \learningrate_t \avggd_t + \learningrate_t \avggd_t||^2 \ka
			&= \underbrace{|| \avgweights_t - \weights^* - \learningrate_t \hspace{-0.5mm} \avggd_t||^2}_{A_1} \ka
			&\quad +  2\learningrate_t \underbrace{\langle \avgweights_t \hspace{-0.5mm} - \hspace{-0.5mm} \weights^* \hspace{-0.5mm} - \hspace{-0.5mm} \learningrate_t \avggd_t, \avggd_t \hspace{-0.5mm} - \hspace{-0.5mm} \avgsgd \rangle}_{A_2} \hspace{-0.5mm} + \hspace{-0.5mm} \underbrace{\learningrate_t^2 ||\avgsgd_t  \hspace{-0.5mm} -  \hspace{-0.5mm} \avggd_t||^2}_{A_3}. \label{SGD_LEMMA_1}
		\end{align} 
		Since $\E[\avgsgd_t] = \avggd_t$, we know that $A_2$ becomes zero after taking expectation. We also split $A_1$ into the three terms as follows:
		\begin{align}
			A_1 &= || \avgweights_t - \weights^* - \learningrate_t \avggd_t||^2 \ka
			& = ||\avgweights_t - \weights^*||^2 \underbrace{-2 \learningrate_t \langle \avgweights_t - \weights^*, \avggd_t \rangle}_{B_1} + \underbrace{\learningrate_t^2  ||\avggd_t||^2}_{B_2}.
		\end{align}
		We now derive an upper bound of $B_1$. From the definition of $\avgweights_t$ and $\avggd_t$, we express $B_1$ as
		\begin{align}
			B_1 & \hspace{-0.5mm} = \hspace{-0.5mm} -2 \learningrate_t \langle \avgweights_t \hspace{-0.5mm} - \hspace{-0.5mm} \weights^*, \avggd_t \rangle = \hspace{-0.5mm} -2 \learningrate_t \hspace{-0.75mm} \sum_{k=1}^{\Whole} \prob \langle \avgweights_t \hspace{-0.5mm} -  \hspace{-0.5mm} \weights^*, \nabla \globall_k (\qweights_t) \rangle \ka
			&= -2 \learningrate_t \sum_{k=1}^{\Whole} \prob \langle \avgweights_t - \qweights_t, \nabla \globall_k (\qweights_t) \rangle \ka
			&\quad -2 \learningrate_t \sum_{k=1}^{\Whole} \prob \langle \qweights_t - \weights^*, \nabla \globall_k (\qweights_t) \rangle.
		\end{align}
		We first derive an upper bound of $-\langle \avgweights_t - \qweights_t, \nabla \globall_k (\qweights_t) \rangle$ using the Cauchy-Schwarz inequality as well as arithmetic mean and geometric mean inequalities as follows:
		\begin{align}
			&- \hspace{-0.2mm} \langle \avgweights_t \hspace{-0.5mm}- \hspace{-0.5mm}  \qweights_t \hspace{-0.8mm} , \hspace{-0.5mm} \nabla \globall_k(\qweights_t) \rangle \ka 
			& \hspace{-0.5mm} \leq \hspace{-0.5mm} \frac{1}{\sqrt{\learningrate_t}} ||\qweights_t \hspace{-1.5mm} - \hspace{-0.5mm} \avgweights_t|| \sqrt{\learningrate_t} ||\nabla \globall_k(\qweights_t)|| \ka
			&\leq \frac{1}{2\learningrate_t} ||\qweights_t - \avgweights_t||^2 + \frac{\learningrate_t}{2} ||\nabla \globall_k(\qweights_t)||^2. \label{A_1_1}
		\end{align}
		We use the assumption of $\strong$-convexity of the loss function to derive an upper bound of $-\langle \qweights_t - \weights^*, \nabla \globall_k (\qweights_t) \rangle $. From the fact that $\globall_k(\weights^*) \geq \globall_k(\qweights_t) + \langle \weights^* -\qweights_t, \nabla \globall_k(\qweights_t) \rangle + \frac{\strong}{2} ||\weights^* - \qweights_t||^2$, we have
		\begin{align}
			-\langle \qweights_t - \weights^*, \nabla \globall_k (\qweights_t) \rangle &\leq -\{\globall_k(\qweights_t) - \globall_k(\weights^*)\} \ka
			&\quad -\frac{\strong}{2}||\weights^* - \qweights_t||^2. \label{A_1_2}
		\end{align}
		For $B_2$, we use $\smooth$-smoothness of the loss function to obtain the upper bound as below
		\begin{align}
			B_2 = \learningrate_t^2 ||\avggd_t||^2 &\leq \learningrate_t^2  \sum_{k=1}^{\Whole} \prob ||\nabla \globall_k(\qweights_t)||^2 \ka
			& \leq 2\smooth\learningrate_t^2 \sum_{k=1}^{\Whole} \prob (\globall_k(\qweights_t) - \globall_k^*). \label{B_2}
		\end{align}
		Then, we obtain an upper bound of $A_1$ using \eqref{A_1_1}, \eqref{A_1_2}, and \eqref{B_2} as follows
		\begin{align}
			A_1 & = \hspace{-0.5mm} ||\avgweights_t - \weights^*||^2 - \strong\learningrate_t \sum_{k=1}^{\Whole} \prob ||\qweights_t - \weights^*||^2 \ka
			&\quad + \sum_{k=1}^{\Whole} \prob ||\qweights_t - \avgweights_t||^2 + \learningrate_t^2 \sum_{k=1}^{\Whole} \prob ||\nabla \globall_k(\qweights_t)||^2 \ka
			&\quad -2\learningrate_t \sum_{k=1}^{\Whole} \prob 
			\left\{ \globall_k(\qweights_t) - \globall_k(\weights^*)
			\right\}
			\ka
			&\quad + 2\smooth\learningrate_t^2 \sum_{k=1}^{\Whole} \prob 
			\left\{
			\globall_k(\qweights_t) - \globall_k^* 
			\right\} \ka
			&\leq \hspace{-0.5mm} ||\avgweights_t \hspace{-0.0mm} -  \hspace{-0.0mm} \weights^*||^2 \hspace{-0.0mm} - \hspace{-0.0mm} \strong\learningrate_t \hspace{-0.0mm} \sum_{k=1}^{\Whole} \prob \hspace{-0.0mm} ||\qweights_t \hspace{-0.0mm} - \hspace{-0.0mm} \weights^*||^2  \ka
			&\quad + \hspace{-0.8mm} \rho \learningrate_t \hspace{-0.7mm} \sum_{k=1}^{\Whole} \prob \hspace{-0.0mm} ||\qweights_t \hspace{-1.5mm} - \hspace{-0.5mm} \avgweights_t||^2 \hspace{-0.9mm} + \hspace{-0.9mm}
			\underbrace{4\smooth\learningrate_t^2  \hspace{-0.8mm} \sum_{k=1}^{\Whole} \prob \hspace{-0.5mm}
			\left\{
			\hspace{-0.7mm} \globall_k(\qweights_t \hspace{-0.2mm} ) \hspace{-0.5mm} - \hspace{-0.7mm} \globall_k^* \hspace{-0.7mm}
			\right\}}_{C} \ka
			&\quad \underbrace{-2\learningrate_t \sum_{k=1}^{\Whole} \prob
			\left\{
			\globall_k(\qweights_t) - \globall_k(\weights^*)
			\right\}}_{C},
		\end{align}
		where the last inequality follows from the $\smooth$-smoothness of the loss function using $||\nabla\globall_k(\qweights_t)||^2 \leq 2\smooth (\globall_k(\qweights_t) - \globall_k^*)$ and $\smart \learningrate_t \geq 1$ with $\smart \gg 1$. Note that $\globall_k^*$ is the minimum value of $\globall_k$. For $\learningrate_t \leq \frac{1}{2\smooth}$, we can derive the upper bound of $C$ as follows 
		\begin{align}
			C &\leq \hspace{-0.0mm} 4\smooth\learningrate_t^2 \hspace{-0.0mm} \sum_{k=1}^{\Whole} \prob \hspace{-0.0mm}
			\left\{
			\globall_k (\qweights_t) \hspace{-0.0mm} - \hspace{-0.0mm} \globall_k^*  \hspace{-0.0mm} - \hspace{-0.0mm}  \globall_k (\qweights_t)  \hspace{-0.0mm} + \hspace{-0.0mm} \globall_k(\weights^*) \hspace{-0.0mm}
			\right\} \ka
			&= 4\smooth\learningrate_t^2 \sum_{k=1}^{\Whole} \prob \Noniid = 4\smooth \learningrate_t^2 \Noniid.
		\end{align}
		Then, $A_1$ can be upper bounded as below
		\begin{align}
			A_1 &\leq ||\avgweights_t - \weights^*||^2 - \strong\learningrate_t \sum_{k=1}^{\Whole} \prob ||\qweights_t - \weights^*||^2 \ka
			&\quad + \rho \learningrate_t \hspace{-0.0mm} \sum_{k=1}^{\Whole} \prob \hspace{-0.0mm} ||\qweights_t - \hspace{-0.0mm} \avgweights_t||^2+ 4\learningrate_t^2\smooth \Noniid.
		\end{align}
		Next, we derive $||\qweights_t - \weights^*||^2$ in $A_1$ as follows
		\begin{align}
			||\qweights_t - \weights^*||^2 &= ||\qweights_t - \weights^k_t + \weights^k_t - \weights^*||^2 \ka
			&= ||\qweights_t - \weights^k_t||^2 + ||\weights^k_t - \weights^*||^2 \ka
			&\quad + 2 \langle \qweights_t - \weights^k_t, \weights^k_t - \weights^* \rangle.
		\end{align}
		Note that $\langle \qweights_t - \weights^k_t, \weights^k_t - \weights^* \rangle$ becomes zero after taking expectation due to Lemma \ref{Lem1}. Then, we can bound $A_1$ as follows
		\begin{align}
			A_1 &\leq (1-\strong \learningrate_t) ||\avgweights_t - \weights^*||^2 - \strong \learningrate_t \sum_{k=1}^{\Whole} \prob ||\qweights_t - \weights^k_t||^2 \ka
			&\quad +\smart \learningrate_t \hspace{-0.0mm} \sum_{k=1}^{\Whole} \hspace{-0.0mm} \prob ||\qweights_t \hspace{-0.0mm} - \hspace{-0.0mm} \avgweights_t||^2  + 4\smooth \learningrate_t^2 \Noniid \label{A1_last}
		\end{align}
		Now we obtain the expectation of \eqref{SGD_LEMMA_1} using \eqref{A1_last} as follows
		\begin{align}
			&\E  \hspace{-0.0mm} \left[
			||\avgv_{t+1} \hspace{-0.0mm} - \hspace{-0mm} \weights^*||^2
			\right]\ka
			&\hspace{-0.0mm} \leq \hspace{-0.0mm} (1 \hspace{-0.0mm} - \hspace{-0.0mm}\strong \learningrate_t) \E \hspace{-0.0mm}
			\left[
			||\avgweights_t \hspace{-0.0mm} - \hspace{-0.0mm} \weights^*||^2
			\right] \hspace{-0.0mm}
			+ \hspace{-0.0mm} \learningrate_t^2 \E \hspace{-0.0mm} \left[ 
			||\avgsgd_t \hspace{-0.0mm} - \hspace{-0.0mm} \avggd_t||^2
			\right] \ka
			& \quad- \strong \learningrate_t \sum_{k=1}^{\Whole} \prob \E \left[
			||\qweights_t - \weights^k_t||^2
			\right] \ka
			& \quad  +  \smart \learningrate_t \sum_{k=1}^{\Whole} \prob  \E  \left[
			 ||\avgweights_t  - \qweights_t||^2 
			\right] + 4\smooth \learningrate_t^2 \Noniid \label{SGD_result_2}
		\end{align}
		To further bound \eqref{SGD_result_2}, we express $\E \left[ 
		||\avgsgd_t - \avggd_t||^2
		\right]$ as 
		\begin{align}
			\E \hspace{-1mm} \left[ 
			||\avgsgd_t \hspace{-0.7mm} - \hspace{-0.7mm} \avggd_t||^2
			\right]
			&= \hspace{-0.7mm} \sum_{k=1}^{\Whole} \prob^2  \E \hspace{-0.7mm} \left[ \hspace{-0.5mm} \bigg|\bigg|
			\nabla \loss(\qweights_t \hspace{-0.5mm} , \minibatch_t^k) \hspace{-0.2mm} - \hspace{-0.2mm} \nabla \loss(\qweights_t)
			\bigg|\bigg|^2 \right] \label{sgd_variance_bound_in_proof} \ka
			&\leq  \sum_{k=1}^{\Whole} \prob^2 \sgdvariance_k^2,
		\end{align}
		where \eqref{sgd_variance_bound_in_proof} is from $\E[\nabla \loss(\qweights_t, \minibatch_t^k)] = \nabla \loss(\qweights_t)$ and the last inequality is from Assumption \ref{assumption 1}. We also derive the upper bound of $\E \left[ ||\avgweights_t - \qweights_t||^2 \right]$ as below
		\begin{align}
			\E \left[ ||\avgweights_t - \qweights_t||^2 \right] &= \E \left[
			||\weights^k_t - \qweights_t||^2 + ||\avgweights_t - \weights^k_t||^2 \right. \ka
			&\left. \quad + 2 \langle \weights^k_t - \qweights_t, \avgweights_t - \weights^k_t \rangle
			\right] \ka
			&\leq \E \left[
			||\weights^k_t - \qweights_t||^2 
			\right] 
			+ 4\learningrate_t^2 (\SGDrun-1)^2 \sgdbound^2,
		\end{align}
		where the last inequality is from Lemma \ref{Lem1} and the result of \cite{Xi:20} for $\learningrate_t \leq 2\learningrate_{t+\SGDrun}$ using
		\begin{equation} \sum_{k=1}^{\Whole} \prob \E \left[ || \avgweights_t - \weights^k_t||^2 \right] \leq 4\learningrate_t^2 (\SGDrun-1)^2 \sgdbound^2.
		\end{equation}
		Then, we can obtain Lemma \ref{Lem2} by using \eqref{quant_variance} in Lemma \ref{Lem1}.
	\end{proof}
	
	\subsection{Proof of Theorem \ref{Thm1}} \label{proof_Thm1}
	Since we use quantization in both local training and transmission, we cannot directly use the result of \cite{Xi:20} to derive the convergence rate due to the quantization errors. We first define an additional auxiliary variable as done in \cite{SZ:21} to prove Theorem \ref{Thm1} as below 
	\begin{align}
		\uweights^k_{t+1} = \begin{cases}
			\vweights^k_{t+1}  \quad &\text{if} \quad t+1 \not\in \globalset, \\
			\frac{1}{\schedulesize} \sum_{k \in \mathcal{N}_{t+1}} \vweights^k_{t+1} \quad &\text{if} \quad t+1 \in \globalset.
		\end{cases}
	\end{align}
	We also define $\avgu_t =\sum_{k=1}^{\Whole} \prob \uweights^k_t$ for convenience. Since we are interested in the result of global iterations, we focus on $t+1 \in \globalset$. Then, we have
	\begin{align}
		||\avgweights_{t+1} - \weights^*||^2 
		&= \underbrace{||\avgweights_{t+1} - \avgu_{t+1}||^2 }_{D_1} + \underbrace{||\avgu_{t+1} - \weights^*||^2}_{D_2} \ka
		&\quad  + \underbrace{ 2 \langle \avgweights_{t+1} - \avgu_{t+1}, \avgu_{t+1} - \weights^* \rangle}_{D_3}. \label{D1}
	\end{align}
	To simplify  \eqref{D1}, we adopt the result of $\avgweights_{t+1}$ and $\avgu_{t+1}$ from \cite{SZ:21} as follows:
	\begin{align}
		&\E[\avgweights_{t+1}] = \avgu_{t+1}, \label{Thm 1-1} \\
		&\E \left[ || \avgweights_{t+1} - \avgu_{t+1} ||^2 \right] \leq \frac{4 \paramsize \learningrate_t^2 \SGDrun \sgdbound^2}{\schedulesize 2^{2\tprecision}}. \label{Thm 1-2}
	\end{align}
	Then, we can know that $D_3$ becomes zero after taking the expectation from \eqref{Thm 1-1} and $D_1$ can be bounded by \eqref{Thm 1-2}. We further obtain the upper bound $D_2$ as below
	\begin{align}
		D_2 
		&= \underbrace{||\avgu_{t+1} - \avgv_{t+1}||^2}_{E_1} + \underbrace{||\avgv_{t+1} - \weights^*||^2}_{E_2} \ka
		&\quad + \underbrace{2 \langle \avgu_{t+1} - \avgv_{t+1}, \avgv_{t+1} - \weights^* \rangle}_{E_3}. \label{E3}
	\end{align}
	We leverage the result of the random scheduling from \cite{SZ:21} to simplify  \eqref{E3} as follows
	\begin{align}
		&\E[\avgu_{t+1}] = \avgv_{t+1} \label{Thm 1-3} \\
		&\E[\avgv_{t+1} - \avgu_{t+1}||^2] \leq \frac{4}{\schedulesize} \learningrate_t^2 \SGDrun^2 \sgdbound^2.  \label{Thm 1-4}
	\end{align}
	We can see that $E_3$ will vanish due to \eqref{Thm 1-3}. $E_1$ and $E_2$ can be upper bounded by \eqref{Thm 1-4} and Lemma \ref{Lem2}, respectively. Therefore, we have 
	\begin{align}
		\E 
		\left[ 
		|| \avgweights_{t+1}  -   \weights^*||
		\right]^2 
		&\leq  \E  
		\left[ 
		|| \avgv_{t+1} \hspace{-0.5mm} - \hspace{-0.5mm} \weights^* ||
		\right] 
		\hspace{-0.mm} +  \hspace{-0.mm} \frac{4\learningrate_t^2 \sgdbound^2}{\schedulesize} \hspace{-0.mm}
		\left( \hspace{-0.mm}
		\frac{\paramsize \SGDrun}{2^{2\tprecision}} \hspace{-0.mm}  + \hspace{-0.mm} \SGDrun^2 \hspace{-0.mm}
		\right) \ka
		&\leq \hspace{-0.5mm} (1\hspace{-0.7mm} - \hspace{-0.7mm} \strong \learningrate_t) \E \hspace{-0.8mm} \left[ 
		|| \avgweights_t  \hspace{-0.7mm} - \hspace{-0.7mm} \weights^*||^2 
		\right]
		\hspace{-0.7mm} + \hspace{-0.7mm} \learningrate_t^2 \longconsttwo  \hspace{-0.5mm} + \learningrate_t \longconst  \label{Thm1-5}, 
	\end{align}
	where 
	\begin{align}
		&\longconst = \frac{\paramsize(\smart - \strong)}{2^{2\precision}}, \ka
		& \longconsttwo = \sum_{k=1}^{\Whole} \prob^2 \sgdvariance_k^2 + 4(\SGDrun-1)^2\sgdbound^2 + \frac{4\paramsize\SGDrun\sgdbound^2}{\schedulesize 2^{2\tprecision}} + \frac{4\SGDrun^2 \sgdbound^2}{\schedulesize} + 4\smooth \Noniid. \label{Thm1-6}
	\end{align}
Since $ \E \left[ || \avgweights_{t} - \weights^*|| \right] \leq \frac{\beta^2 \longconsttwo}{(\beta \mu - 1)(t+\gamma)} + \frac{\beta \longconst}{\beta\strong-1}$ satisfies \eqref{Thm1-6} for $ \learningrate_t = \frac{\beta}{t+\gamma}$  as shown in \cite{Xi:20}. Then, we can obtain Theorem \ref{Thm1} from $\smooth$ - smoothness of the loss function using $\E[\globall(\avgweights_{t+1}) - \globall(\weights^*)] \leq \frac{\smooth}{2} 	\E \left[ || \avgweights_{t+1} - \weights^*|| \right]^2$. Finally, we change the time scale to local iteration.
\end{appendix}

	\bibliographystyle{IEEEtran}
	\bibliography{Bibtex/StringDefinitions,Bibtex/IEEEabrv,Bibtex/mybib}

\begin{thebibliography}{10}
\providecommand{\url}[1]{#1}
\csname url@samestyle\endcsname
\providecommand{\newblock}{\relax}
\providecommand{\bibinfo}[2]{#2}
\providecommand{\BIBentrySTDinterwordspacing}{\spaceskip=0pt\relax}
\providecommand{\BIBentryALTinterwordstretchfactor}{4}
\providecommand{\BIBentryALTinterwordspacing}{\spaceskip=\fontdimen2\font plus
\BIBentryALTinterwordstretchfactor\fontdimen3\font minus
  \fontdimen4\font\relax}
\providecommand{\BIBforeignlanguage}[2]{{%
\expandafter\ifx\csname l@#1\endcsname\relax
\typeout{** WARNING: IEEEtran.bst: No hyphenation pattern has been}%
\typeout{** loaded for the language `#1'. Using the pattern for}%
\typeout{** the default language instead.}%
\else
\language=\csname l@#1\endcsname
\fi
#2}}
\providecommand{\BIBdecl}{\relax}
\BIBdecl

\bibitem{ICC_version}
M.~Kim, W.~Saad, M.~Mozaffari, and M.~Debbah, ``On the tradeoff between energy,
  precision, and accuracy in federated quantized neural networks,'' in
  \emph{Proc. of IEEE Int. Conf. Commun.}, Seoul, South Korea, May 2022.

\bibitem{CM:21}
M.~Chen, Z.~Yang, W.~Saad, C.~Yin, H.~V. Poor, and S.~Cui, ``A joint learning
  and communications framework for federated learning over wireless networks,''
  \emph{{IEEE} Trans. Wireless Commun.}, vol.~20, no.~1, pp. 269--283, Jan.
  2021.

\bibitem{Ro:19}
R.~Schwartz, J.~Dodge, N.~A. Smith, and O.~Etzioni, ``Green {AI},'' \emph{arXiv
  preprint arXiv:1907.10597}, 2019.

\bibitem{HB:16}
H.~B. McMahan, E.~Moore, D.~Ramage, S.~Hampson, and B.~A. Arcas,
  ``Communication-efficient learning of deep networks from decentralized
  data,'' \emph{arXiv preprint arXiv:1602.05629}, 2017.

\bibitem{MB:17}
B.~Moons, K.~Goetschalckx, N.~Van~Berckelaer, and M.~Verhelst, ``Minimum energy
  quantized neural networks,'' in \emph{Proc. of Asilomar Conf. on Signals,
  Systems, and Computers}, Pacific Grove, CA, USA, Apr. 2017.

\bibitem{Stesa:21}
S.~Savazzi, V.~Rampa, S.~Kianoush, and M.~Bennis, ``An energy and carbon
  footprint analysis of distributed and federated learning,'' \emph{arXiv
  preprint arXiv:2206.10380}, 2022.

\bibitem{Ngutra:19}
N.~H. Tran, W.~Bao, A.~Zomaya, M.~N.~H. Nguyen, and C.~S. Hong, ``Federated
  learning over wireless networks: Optimization model design and analysis,'' in
  \emph{Proc. of IEEE Conf. on Computer Commun.}, Paris, France, May 2019.

\bibitem{Zhaya:21}
Z.~Yang, M.~Chen, W.~Saad, C.~S. Hong, and M.~Shikh-Bahaei, ``Energy efficient
  federated learning over wireless communication networks,'' \emph{{IEEE}
  Trans. Wireless Commun.}, vol.~20, no.~3, pp. 1935--1949, Mar. 2021.

\bibitem{Khaung:21}
Q.~Zeng, Y.~Du, K.~Huang, and K.~K. Leung, ``Energy-efficient resource
  management for federated edge learning with cpu-gpu heterogeneous
  computing,'' \emph{{IEEE} Trans. Wireless Commun.}, vol.~20, no.~12, pp.
  7947--7962, Dec. 2021.

\bibitem{BL:21}
B.~Luo, X.~Li, S.~Wang, J.~Huangy, and L.~Tassiulas, ``Cost-effective federated
  learning design,'' in \emph{Proc. of IEEE Conf. on Computer Commun.},
  Vancouver, BC, Canada, May 2021.

\bibitem{BA:20}
R.~Balakrishnan, M.~Akdeniz, S.~Dhakal, and N.~Himayat, ``Resource management
  and fairness for federated learning over wireless edge networks,'' in
  \emph{Proc. of IEEE Workshop on Signal Process. Advances in Wireless
  Commun.}, Atlanta, GA, USA, May 2020.

\bibitem{additional1}
G.~Zhu, Y.~Du, D.~Gündüz, and K.~Huang, ``One-bit over-the-air aggregation
  for communication-efficient federated edge learning: Design and convergence
  analysis,'' \emph{{IEEE} Trans. Wireless Commun.}, vol.~20, no.~3, pp.
  2120--2135, Mar. 2021.

\bibitem{addtional2}
P.~Liu, J.~Jiang, G.~Zhu, L.~Cheng, W.~Jiang, W.~Luo, Y.~Du, and Z.~Wang,
  ``Training time minimization for federated edge learning with optimized
  gradient quantization and bandwidth allocation,'' \emph{Frontiers of
  Information Technology \& Electronic Engineering}, vol.~23, no.~8, pp.
  1247--1263, 2022 \color{black}.

\bibitem{FE:21}
C.~Feng, Z.~Zhao, Y.~Wang, T.~Q. Quek, and M.~Peng, ``On the design of
  federated learning in the mobile edge computing systems,'' \emph{{IEEE}
  Trans. Commun.}, vol.~69, no.~9, pp. 5902--5916, Sep. 2021.

\bibitem{Ch:22}
R.~Chen, L.~Li, K.~Xue, C.~Zhang, M.~Pan, and Y.~Fang, ``Energy efficient
  federated learning over heterogeneous mobile devices via joint design of
  weight quantization and wireless transmission,'' \emph{{IEEE} Trans. Mobile
  Comput.}, pp. 1--13, Oct. 2022.

\bibitem{IH:16}
I.~Hubara, M.~Courbariaux, D.~Soudry, R.~El-Yaniv, and Y.~Bengio, ``Quantized
  neural networks: Training neural networks with low precision weights and
  activations.'' \emph{arXiv preprint arXiv:1609.07061}, 2016.

\bibitem{SG:15}
S.~Gupta, A.~Agrawal, K.~Gopalakrishnan, and P.~Narayanan, ``Deep learning with
  limited numerical precision,'' in \emph{Proc. of International Conference on
  Machine Learning (ICML)}, Lille, France, Jul. 2015.

\bibitem{DOREFA}
S.~Zhou, Y.~Wu, Z.~Ni, X.~Zhou, H.~Wen, and Y.~Zou, ``{DoReFa-Net}: Training
  low bitwidth convolutional neural networks with low bitwidth gradients,''
  \emph{arXiv preprint arXiv:1606.06160}, 2018.

\bibitem{SZ:21}
S.~Zheng, C.~Shen, and X.~Chen, ``Design and analysis of uplink and downlink
  communications for federated learning,'' \emph{{IEEE} J. Sel. Areas Commun.},
  vol.~39, no.~7, Jul. 2021.

\bibitem{MB:18}
B.~Moons, D.~Bankman, and M.~Verhelst, \emph{Embedded Deep Learning,
  Algorithms, Architectures and Circuits for Always-on Neural Network
  Processing}.\hskip 1em plus 0.5em minus 0.4em\relax Springer, 2018.

\bibitem{Ev:20}
U.~Evci, T.~Gale, J.~Menick, P.~S. Castro, and E.~Elsen, ``Rigging the lottery:
  Making all tickets winners,'' in \emph{Proc. of International Conference on
  Machine Learning (ICML)}, Vienna, Austria, Apr. 2020, pp. 2943--2952.

\bibitem{BJ:14}
E.~Bjornson, E.~A. Jorswieck, M.~Debbah, and B.~Ottersten, ``Multiobjective
  signal processing optimization: The way to balance conflicting metrics in 5g
  systems,'' \emph{IEEE Signal Processing Magazine}, vol.~31, no.~6, pp.
  14--23, Nov. 2014.

\bibitem{Xi:20}
X.~Li, K.~Huang, W.~Yang, S.~Wang, and Z.~Zhang, ``On the convergence of fedavg
  on non-iid data,'' in \emph{Proc. of International Conference on Learning
  Representations (ICLR)}, May 2020.

\bibitem{ZY:13}
Z.~Yuchen, D.~J. C., and W.~M. J., ``Communication-efficient algorithms for
  statistical optimization,'' \emph{J. Mach. Learn. Res.}, vol.~14, no.~1, p.
  3321–3363, Jan. 2013.

\bibitem{NBI}
I.~Das and J.~E. Dennis, ``Normal-boundary intersection: A new method for
  generating the pareto surface in nonlinear multicriteria optimization
  problems,'' \emph{SIAM journal on optimization}, vol.~8, no.~3, pp. 631--657,
  Aug. 1998.

\bibitem{Cardano}
R.~Witu{\l}a and D.~S{\l}ota, ``Cardano's formula, square roots, chebyshev
  polynomials and radicals,'' \emph{Journal of Mathematical Analysis and
  Applications}, vol. 363, no.~2, pp. 639--647, Feb. 2010.

\bibitem{Nonlinearprogramming}
D.~P. Bertsekas, ``Nonlinear programming,'' \emph{Journal of the Operational
  Research Society}, vol.~48, no.~3, pp. 334--334, 1997.

\bibitem{Convex}
S.~Boyd, S.~P. Boyd, and L.~Vandenberghe, \emph{Convex optimization}.\hskip 1em
  plus 0.5em minus 0.4em\relax Cambridge university press, 2004.

\bibitem{gamethoerybook}
Z.~Han, D.~Niyato, W.~Saad, T.~Başar, and A.~Hjørungnes, \emph{Game Theory in
  Wireless and Communication Networks: Theory, Models, and Applications}.\hskip
  1em plus 0.5em minus 0.4em\relax Cambridge University Press, 2011.

\bibitem{LE:08}
E.~Larsson and E.~Jorswieck, ``Competition versus cooperation on the miso
  interference channel,'' \emph{{IEEE} J. Sel. Areas Commun.}, vol.~26, no.~7,
  pp. 1059--1069, Sep. 2008.

\bibitem{PH:07}
P.~Hyunggon and M.~van~der Schaar, ``Bargaining strategies for networked
  multimedia resource management,'' \emph{{IEEE} Trans. Signal Process.},
  vol.~55, no.~7, pp. 3496--3511, Jul. 2007.

\bibitem{YE:21}
R.~Yedida, S.~Saha, and T.~Prashanth, ``Lipschitzlr: Using theoretically
  computed adaptive learning rates for fast convergence,'' \emph{Applied
  Intelligence}, vol.~51, Mar. 2021.

\bibitem{HO:15}
T.~Hofmann, A.~Lucchi, S.~Lacoste-Julien, and B.~McWilliams, ``Variance reduced
  stochastic gradient descent with neighbors,'' in \emph{Proc. of Neural
  Information Processing Systems (NeurIPS)}, Montreal, Canada, Dec. 2015.

\bibitem{convexity}
Q.~Jin and A.~Mokhtari, ``Exploiting local convergence of quasi-newton methods
  globally: Adaptive sample size approach,'' in \emph{Proc. of Neural
  Information Processing Systems (NeurIPS)}, Virtual, Dec. 2021.

\bibitem{OL:15}
A.~Øland and B.~Raj, ``Reducing communication overhead in distributed learning
  by an order of magnitude (almost),'' in \emph{Proc. of IEEE Int. Conf.
  Acoustics, Speech, and Signal Processing}, South Brisbane, QLD, Australia,
  2015, pp. 2219--2223.

\bibitem{SA:20}
Y.~Sarikaya and O.~Ercetin, ``Motivating workers in federated learning: A
  stackelberg game perspective,'' \emph{{IEEE} Net. Lett.}, vol.~2, no.~1, pp.
  23--27, Oct. 2020.

\bibitem{RE:20}
A.~Reisizadeh, A.~Mokhtari, H.~Hassani, A.~Jadbabaie, and R.~Pedarsani,
  ``{FedPAQ}: A communication-efficient federated learning method with periodic
  averaging and quantization,'' in \emph{Proc. of International Conference on
  Artificial Intelligence and Statistics (AISTATS)}, Virtual Conference, Jun.
  2020.

\end{thebibliography}

\end{document}